\pdfoutput=1
\documentclass[11pt]{article}

\usepackage{acl}

\usepackage{times}
\usepackage{latexsym}
\usepackage{subcaption}
\usepackage{multirow}
\usepackage{graphicx}
\usepackage{siunitx}
\usepackage{makecell}
\usepackage{tabularx}
\usepackage{float}

\usepackage[T1]{fontenc}
\usepackage[utf8]{inputenc}

\usepackage{microtype}

\newcommand{\tr}[1]{\textcolor{red}{#1}}

\title{MM-Claims: A Dataset for Multimodal Claim Detection in Social Media}

 \author{Gullal S. Cheema$^1$, Sherzod Hakimov$^{1,3}$, Abdul Sittar$^2$, \\ {\bf Eric M{\"u}ller-Budack$^{1,3}$, Christian Otto$^3$, and Ralph Ewerth$^{1,3}$} \\
	$^1$TIB -- Leibniz Information Centre for Science and Technology, Hannover, Germany \\
	$^2$Jozef Stefan Institute, Ljubljana, Slovenia \\ 
	$^3$L3S Research Center, Leibniz University Hannover, Germany \\
	\texttt{\{gullal.cheema, sherzod.hakimov, eric.mueller\}@tib.eu} \\
	\texttt{\{christian.otto, ralph.ewerth\}@tib.eu} \\
	\texttt{abdul.sittar@ijs.si}}

\begin{document}
\maketitle
%% article.
\begin{abstract}
In recent years, the problem of misinformation on the web has become widespread across languages, countries, and various social media platforms. Although there has been much work on automated fake news detection, the role of images and their variety are not well explored. In this paper, we investigate the roles of image and text at an earlier stage of the fake news detection pipeline, called claim detection. For this purpose, we introduce a novel dataset, \emph{MM-Claims}, which consists of tweets and corresponding images over three topics: \emph{COVID-19}, \emph{Climate Change} and broadly \emph{Technology}. The dataset contains roughly \num{86000} tweets, out of which \num{3400} are labeled manually by multiple annotators for the training and evaluation of multimodal models. We describe the dataset in detail, evaluate strong unimodal and multimodal baselines, and analyze the potential and drawbacks of current models.
\end{abstract}

\section{Introduction}\label{sec:intro}
The importance of combating misinformation was once again illustrated by the coronavirus pandemic, which came along with a lot of "potentially lethal" misinformation.
At the beginning of the COVID-19 pandemic, the United Nations (UN)~\cite{UN_link} started even using the term ``infodemic'' for this phenomenon of misinformation and called for proper dissemination of reliable facts. 
However, tackling misinformation online and specifically on social media platforms is  challenging due to the variety of information, volume, and speed of streaming data. 
As a consequence, several studies have explored different aspects of COVID-19 misinformation online including sharing patterns~\cite{pennycook2020fighting}, platform-dependent engagement patterns~\cite{cinelli2020covid}, web search behaviors~\cite{rovetta2020covid}, and fake images~\cite{sanchez2020fake}. 

\begin{figure}[t]
	\centering
	\includegraphics[width=\linewidth]{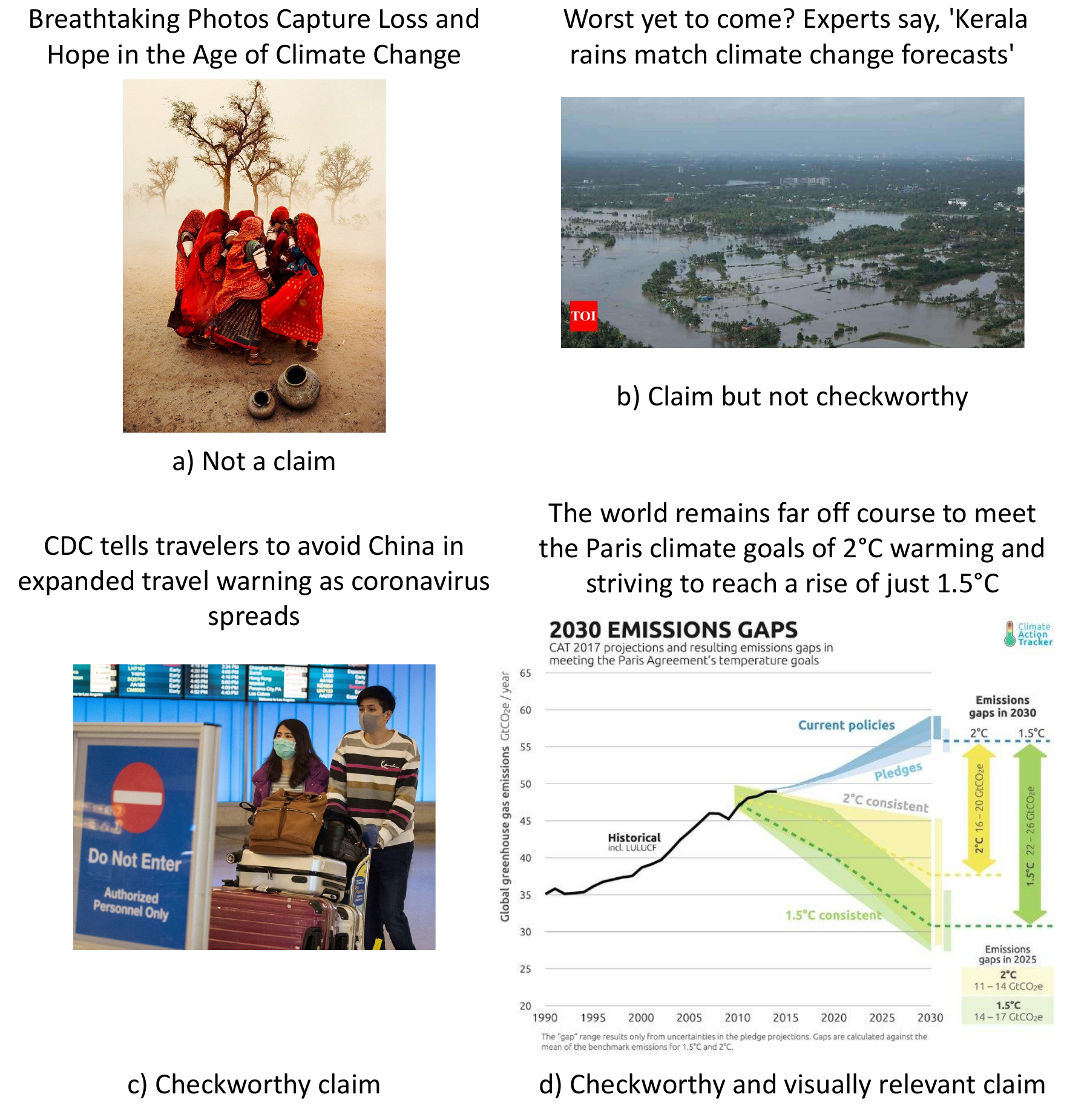}
	\caption{Examples for each of the four classes in the MM-Claims dataset: a) \textbf{not a claim} (both image and text together abstractly represent effects of climate change), b) \textbf{claim but not checkworthy} (claim in text, but lacks details like to which experts is referred to, while image is relevant), c) \textbf{checkworthy} but not visually relevant (claim in text targets CDC and China but the image is a stock photograph), and d) \textbf{checkworthy and visually relevant} (claim in text and in image with important details in both).}
	\label{fig:intro_ex}
\end{figure}

%In this work, 
We are primarily interested in claims on social media from a multimodal perspective~(Figure~\ref{fig:intro_ex}). 
Claim detection can be seen as an initial step in fighting misinformation and as a precursor to prioritize potentially false information for fact-checking. 
Traditionally, claim detection is studied from a linguistic standpoint where both syntax~\cite{DBLP:conf/semco/RosenthalM12} and semantics~\cite{levy2014context} of the language matter to detect a claim accurately. However, claims or fake news on social media are not bound to just one modality and become a complex problem with additional modalities like images and videos. While it is clear that a claim in the text is denoted in verbal form, it can also be part of the visual content or as overlaid text in the image. 
Even though much effort has been spent on the curation of datasets~\cite{DBLP:conf/mediaeval/BoididouPDBRMPK16,nakamura2020fakeddit,DBLP:conf/aaai/JindalS0V020} and the development of computational models for multimodal fake news detection on social media~\cite{ajao2018fake,wang2018eann,khattar2019mvae,DBLP:conf/bigmm/SinghalS0KS19}, hardly any research has focused on multimodal claims~\cite{zlatkova2019fact,cheema2021tib}. 

In this paper, we extend the definitions of claims and check-worthiness from previous work~\cite{barron2020overview,DBLP:conf/ecir/NakovMEBMSAHHBN21} to multimodal claim detection and introduce a novel dataset called \textit{Multimodal Claims (MM-Claims)} curated from Twitter to tackle this critical problem. 
While previous work has focused on factually-verifiable check-worthy~\cite{barron2020overview,DBLP:journals/corr/abs-2005-00033} or general claims (i.e., not necessarily factually-verifiable, e.g.,~\cite{gupta2021lesa}) on a single topic, we focus on three different topics, namely \emph{COVID-19}, \emph{Climate Change} and \emph{Technology}. 

As shown in Figure~\ref{fig:intro_ex}, MM-Claims aims to differentiate between tweets without claims~(Figure~\ref{fig:intro_ex}a) as well as tweets with claims of different types:
\textit{claim but not check-worthy}~(Figure~\ref{fig:intro_ex}b), \textit{check-worthy claim}~(Figure~\ref{fig:intro_ex}c), and \textit{check-worthy visually relevant claim}~(Figure~\ref{fig:intro_ex}d). 

Our contributions can be summarized as follows:
\begin{itemize}
	\setlength\itemsep{0em}
	\item a novel dataset for multimodal claim detection in social media with more than \num{3000} manually annotated and roughly \num{82000} unlabeled image-text tweets is introduced; 
	%The labeled dataset consists of three kinds of annotations that correspond to three different tasks.
	\item we present details about the dataset and the annotation process, %label 
	class definitions, dataset characteristics, and inter-coder agreement;
	\item we provide a detailed experimental evaluation of strong unimodal and multimodal models highlighting the difficulty of the task as well as the role of image and text content.
\end{itemize}

The remainder of the paper is structured as follows. Section~\ref{sec:related} describes the related work on unimodal and multimodal approaches for claim detection. The proposed dataset and the annotation guidelines are presented in Section~\ref{sec:dataset}. We discuss the experimental results of the compared models in Section~\ref{sec:experiments}, while Section~\ref{sec:conclusion} concludes the paper and outlines areas of future work.
\section{Related Work}\label{sec:related}
% The problem of detecting claims in social media has received much attention in recent years because of rising misinformation and the challenge noisy social media text presents. Most of the previous work has focused on processing only textual data. However, the availability of emoticons, video, and images makes it essential to include these types of information in algorithms along with text because users on social media create posts with different modalities and types of information. First, we present previous work on claim detection that utilizes only a single modality, mainly the text content. Next, we proceed with approaches built for multimodal data.

\subsection{Text-based Approaches}
Before research on claim detection targeted social media, pioneering work by \citet{DBLP:conf/semco/RosenthalM12} focused on claims in \emph{Wikipedia} discussion forums. 
They used lexical and syntactic features in addition to sentiment and other statistical features over text. 
Since then, researchers have proposed context-dependent~\cite{levy2014context}, context-independent~\cite{lippi2015context}, cross-domain~\cite{daxenberger2017essence}, and in-domain approaches for claim detection. 
Recently, transformer-based models~\cite{chakrabarty2019imho} have replaced structure-based claim detection approaches due to their success in several Natural Language Processing (NLP) downstream tasks. 
A series of workshops~\cite{barron2020overview,DBLP:conf/ecir/NakovMEBMSAHHBN21} focused on claim detection and verification on Twitter and organized challenges with several sub-tasks on text-based claim detection around the topic of \emph{COVID-19} in multiple languages. 
% The top submissions~\cite{cheema2020checksquare,DBLP:conf/clef/NikolovMKN20,clef-checkthat-williams:2020} used some version of transformer-based models like \emph{BERT}~\cite{devlin2018bert} along with tweet metadata and lexical features. Since the dataset size is typically limited, some approaches used data augmentation~\cite{DBLP:conf/clef/ZenginKK21}, multilingual sentence transformer models~\cite{DBLP:conf/clef/SchlichtPR21} and topic-specific pre-trained transformer models~\cite{DBLP:conf/clef/Martinez-RicoMA21}.
\citet{gupta2021lesa} addressed the limitations of current methods in cross-domain claim detection by proposing a new dataset of about $\sim$\num{10000} claims on \emph{COVID-19}. 
They also proposed a model that combines transformer features with learnable syntactic feature embeddings. 
Another dataset introduced by \citet{DBLP:conf/nldb/IskenderSP021} includes tweets in German about \emph{climate change} for claim and evidence detection. \citet{DBLP:conf/bionlp/WuhrlK21} created a dataset for biomedical Twitter claims related to \emph{COVID-19, measles, cystic fibrosis and depression}. 
One common theme and challenge among all the datasets is the variety of claims where some types of claims (like implicit) are harder to detect than explicit ones where a typical claim structure is present. Table~\ref{tab:datasets} shows a comparison of existing social media based claim datasets, with number of samples, modalities, data sources, language, topic, and type of tasks.

\begin{table*}[]
	\footnotesize
	\begin{tabular}{|l|c|c|c|c|c|c|}
		\hline
		\multicolumn{1}{|c|}{\textbf{Datasets}}             & \textbf{\#Samples}                                                & \textbf{Modality}                                      & \textbf{Data source}                                      & \textbf{Language} & \textbf{Topic}                                                                    & \textbf{Task(s)}                                                                          \\ \hline
		\citet{zlatkova2019fact}*                 & 1233                                                              & \begin{tabular}[c]{@{}c@{}}Image, \\ Text\end{tabular} & \begin{tabular}[c]{@{}c@{}}Snopes,\\ Reuters\end{tabular} & English           & Multi-topic                                                                       & True vs False \\ \hline
		\citet{DBLP:conf/ecir/NakovMEBMSAHHBN21} & $18,014^\dagger$ & Text                                                   & Twitter                                                   & Multi      & Multi-topic$^\dagger$        & \begin{tabular}[c]{@{}c@{}}check-worthiness \\ estimation \end{tabular} 
		\\ \hline &&&&&&\\[-0.5em]
		\citet{gupta2021lesa}                    & 9981                                                              & Text                                                   & Twitter                                                   & English            & COVID-19                                                                          & claim detection                                                                         \\[1.2ex] \hline
		\citet{DBLP:conf/nldb/IskenderSP021}     & 300 pairs                                                         & Text                                                   & Twitter                                                   & German            & Climate change                                                                    & \begin{tabular}[c]{@{}c@{}}claim, evidence\\ detection\end{tabular}                     \\ \hline
		\citet{DBLP:conf/bionlp/WuhrlK21}        & 1200                                                              & Text                                                   & Twitter                                                   & English           & \begin{tabular}[c]{@{}c@{}}Biomedical \&\\ COVID-19\end{tabular}           & \begin{tabular}[c]{@{}c@{}}claim \& \\ claim type \\ detection \end{tabular}                                                                                             \\ \hline
		\textbf{MM-Claims (Ours)}                                 & 3400                                                              & \begin{tabular}[c]{@{}c@{}}Image, \\ Text\end{tabular} & Twitter                                                   & English           & \begin{tabular}[c]{@{}c@{}}COVID-19, \\ Climate Change,\\ Technology\end{tabular} & \begin{tabular}[c]{@{}c@{}}claim, \\ check-worthiness, \\ visual relevance\end{tabular} \\ \hline
	\end{tabular}
	\caption{Comparison of social media based claim datasets. *\citet{zlatkova2019fact} is a mix of actual news photographs (from Reuters) and possibly fake images (from Snopes), which went viral on social media sites like Reddit. $^\dagger$ 1312 samples are in English and only on the topic of COVID-19.}\label{tab:datasets} 
\end{table*}

\subsection{Multimodal Approaches}
From the multimodal perspective, very few works have analyzed the role of images in the context of claims. \citet{zlatkova2019fact} introduced a dataset that consists of claims and is created from the idea of investigating questionable or outright false images which supplement fake news or claims. 
The authors used reverse image search and several image metadata features such as tags from Google Vision API, URL domains and categories, reliability of the image source, etc. 
Similarly, \citet{wang2020fake} performed a large-scale study by analyzing manipulated or misleading images in news discussions on forums like \textit{Reddit}, \textit{4chan} and \textit{Twitter}. 
For claim detection, \citet{DBLP:conf/www/CheemaHME21} extended the text-based claim detection datasets of \citet{barron2020overview} and \citet{gupta2021lesa} with images to evaluate multimodal detection approaches. 
Although previous work has provided multimodal datasets on claims, they are either on veracity (true or false) of claims or labeled only text-based for a single topic (COVID-19).
In terms of multimodal models for image-text data, most previous work is in the related area of multimodal fake news, where several benchmark datasets and models exist for fake news detection~\cite{nakamura2020fakeddit,DBLP:conf/mediaeval/BoididouPDBRMPK16,DBLP:conf/aaai/JindalS0V020} . 
In an early work, \citet{jin2017multimodal} explored rumor detection on Twitter using text, social context (emoticons, URLs, hashtags), and the image by learning a joint representation in a deep recurrent neural network. 
Since then, several improvements have been proposed, such as multi-task learning with an event discriminator~\cite{wang2018eann}, multimodal variational autoencoder~\cite{khattar2019mvae} and multimodal transfer learning using transformers for text and image~\cite{giachanou2020multimodal, DBLP:conf/bigmm/SinghalS0KS19}.
\section{MM-Claims Tasks and Dataset}\label{sec:dataset}
This section describes the problem of multimodal claim detection~(Section~\ref{subsec:task}), the data collection~(Section~\ref{subsec:data_collection}), the guidelines for annotating multimodal claims~(Section~\ref{sec:guideline}), and the annotation process~(Section~\ref{subsec:annotation_process}) to obtain the new dataset.

\subsection{Task Description}\label{subsec:task}
Given a tweet with a corresponding image, the task is to identify important factually-verifiable or check-worthy claims.
In contrast to related work, we introduce a novel dataset for claim detection that is labeled based on both the tweet and the corresponding image, making the task truly multimodal. Our scope of claims is motivated by \citet{DBLP:journals/corr/abs-2005-00033} and \citet{gupta2021lesa}, which have provided detailed annotation guidelines. We restrict our dataset to factually-verifiable claims (as in~\citet{DBLP:journals/corr/abs-2005-00033}) since these are often the claims that need to be prioritized for fact-checking or verification to limit the spread of misinformation. On the other hand, we also include claims that are personal opinions, comments, or claims existing at sub-sentence or sub-clause level (as in~\citet{gupta2021lesa}), with the condition that they are factually-verifiable. Subsequently, we extend the definition of claims to images along with factually-verifiable and check-worthy claims.

\subsection{Data Collection}\label{subsec:data_collection}
In previous work on claim detection in tweets, most of the publicly available English language datasets~\cite{DBLP:journals/corr/abs-2005-00033,barron2020overview,gupta2021lesa,DBLP:conf/ecir/NakovMEBMSAHHBN21} are text-based and on a single topic such as \emph{COVID-19}, or \emph{U.S. 2016 Elections}. To make the problem interesting and broader, we have collected tweets on three topics, \emph{COVID-19}, \emph{Climate Change} and broadly \emph{Technology}, that might be of interest to a wider research community.
Next, we describe the steps for crawling and preprocessing the data.

\subsubsection{Data Crawling}
We have used an existing collection of tweet IDs, where some are topic-specific Twitter dumps, and extracted tweet text and the corresponding image to create a novel multimodal dataset.

\noindent\textbf{COVID-19}: We combined tweets from three Twitter resources \cite{DBLP:journals/corr/abs-2004-03688,DBLP:conf/cikm/DimitrovBF0ZZD20, 781w-ef42-20} that were posted between October 2019 and April 2020. In our dataset, we use tweets in the period from March - April 2020.

\noindent\textbf{Climate Change}: We used a Twitter resource~\cite{DVN/5QCCUU_2019} that contains tweet IDs related to climate change from September 2017 to May 2019. The tweets were originally crawled based on hashtags like \emph{climatechange, climatechangeisreal, actonclimate, globalwarming, climatedeniers, climatechangeisfalse}, etc.

\noindent\textbf{Technology}: For the broad topic of \emph{Technology}, we used the \emph{Tweets\-KB}~\cite{DBLP:conf/esws/FafaliosIND18} corpus.
To avoid the extraction of all the tweets from 2019 to 2020 irrespective of the topic, we followed a two-step process to find tweets remotely related to technology. The corpus is available in form of RDF (Resource Description Framework) triples with attributes like tweet ID, hashtags, entities and emotion labels, but without tweet text or media content details. First, we selected tweet IDs based on hashtags and entities, and only kept those that contain keywords like \emph{technology, cryptocurrency, cybersecurity, machine learning, nano technology, artificial intelligence, IOT, 5G, robotics, blockchain}, etc. The second step of filtering tweets based on a selected set of hashtags for each topic is described in the next subsection.

From the above resources, we collected \num{214715}, \num{28374} and \num{417403} tweets for the topics \emph{COVID-19}, \emph{Climate Change} and \emph{Technology}, respectively. 

\subsubsection{Data Filtering}

We perform a number of filtering steps to remove inconsistent samples: 1) tweets that are not in English or without any text, 2) duplicated tweets based on tweet IDs, processed text and retweets, 3) tweets with corrupted or no images, 4) tweets with images of less than $200 \times 200$ pixels resolution, 5) tweets that have more than six hashtags, and finally, 6) we make a list of the top 300 hashtags in each topic based on count and manually select those related to the selected topics. We only keep those tweets where all hashtags are in the list of top selected hashtags. The hashtags are manually marked because some top hashtags are not relevant to the main topic of interest.
The statistics of tweets after each filtering step are provided in the Appendix (see Table~\ref{tab:stats}). 
In summary, we end up with \num{17771}, \num{4874}, and \num{62887} tweets with images for \emph{COVID-19}, \emph{Climate Change} and \emph{Technology}, respectively.

\subsection{Annotation Guidelines}\label{sec:guideline}

In this section, we provide definitions for all investigated claim aspects, the questions asked to annotators, and the cues and explanations for the annotation questions.
We define a claim as \emph{state or assert that something is the case, typically without providing evidence or proof} using the definition in the Oxford dictionary (like~\citet{gupta2021lesa}). 

The definition of a \emph{factually-verifiable claim} is restricted to claims that can possibly be verified using external sources. These external sources can be reliable websites, books, scientific reports, scientific publications, credible fact-checked news reports, reports from credible organizations like World Health Organization or United Nations. Although we did not provide external links of reliable sources for the content in the tweet, we highlighted named entities that pop-up with the text and image description. External sources are not important at this stage because we are only interested in marking claims, which have possibly incorrect details and information. A list of identifiable cues (extended from~\citet{barron2020overview}) for factually-verifiable claims is provided in the Appendix~\ref{app:claim_definition}.

To define check-worthiness, we follow \citet{barron2020overview} and identify claims as check-worthy if the information in the tweet is, 1) \textit{harmful} (attacks a person, organization, country, group, race, community, etc), or 2) \textit{urgent or breaking news} (news-like statements about prominent people, organizations, countries and events), or 3) \textit{up-to-date} (referring to recent official document with facts, definitions and figures). A detailed description of these cases is provided in the Appendix~\ref{app:claim_definition}. Given these key points, the answer to whether the claim is check-worthy is subjective since it depends on the person's (annotator's) background and knowledge. 
%Contrary to \citet{barron2020overview}, we therefore do not label extra attributes like ``general interest to the public'', ``requiring the attention of professionals'' and ``extent of false information'', which are even more subjective and only provide complementary information.
% to the above discussed check-worthy attributes. 
% Based on these definitions, we decided on the following annotation questions to identify factually-verifiable claims in multimodal data.

% \subsubsection{Annotation Questions}\label{subsec:annotation_questions}
%
\textbf{Annotation Questions}:
Based on the definitions above, we decided on the following annotation questions in order to identify factually-verifiable claims in multimodal data.
% To annotate our dataset, we ask each annotator the following questions, which have to be answered by looking at both image and text in a given tweet:
%
\begin{itemize}
	\setlength\itemsep{0em}
	\item Q1: \emph{Does the image-text pair contain a factually-verifiable claim? - Yes / No} %\\
	% We refer to the definition in Section~\ref{subsec:factual_claim} for identifying factually-verifiable claims.
	\item Q2: \emph{If ``Yes'' to Q1, Does the claim contain harmful, up-to-date, urgent or breaking-news information? - Yes / No} %\\
	% We refer to the definition in Section~\ref{subsec:check_claim} for identifying check-worthy claims.
	\item Q3: \emph{If ``Yes'' to Q1, Does the image contain information about the claim or the claim itself (in the overlaid text)? - Yes / No} %\\%
	% This question intends to identify whether the visual content contributes to a tweet having factually-verifiable claims.  The question is answered ``Yes''  if one of the following cases hold true: 1) there exists a piece of evidence (e.g. an event, action, situation or a person's identity) or illustration of certain aspects in the claim text, or 2) the image contains overlay text that itself contains a claim in a text form.
\end{itemize}
%
%Each annotator was asked to answer these questions by looking at both image and text in a given tweet. 
% 
Question 3~(Q3) intends to identify whether the visual content contributes to a tweet having factually-verifiable claims.  
The question is answered ``Yes'' if one of the following cases hold true: 1)~there exists a piece of evidence (e.g. an event, action, situation or a person's identity) or illustration of certain aspects in the claim text, or 2)~the image contains overlay text that itself contains a claim in a text form.
Please note that we asked the annotators to label tweets with respect to the time they were posted. During our annotation dry runs we observed that there were several false annotations for the tweets where the claims were false but already well known facts.
This aspect intends to ignore the veracity of claims since some of the claims become facts over time. % once they are verified.
In addition, we ignore tweets that are questions and label them as not claims unless the corresponding image consists of a response to the question and is a factually-verifiable claim. 
%More detailed information about the annotation guidelines is provided in Appendix~\ref{sec:annotation_codebook}.

\subsection{Annotation Process}\label{subsec:annotation_process}
Each annotator was asked to answer these questions by looking at both image and text in a given tweet.
We distribute the data among nine external and four expert internal annotators for the annotation of training and evaluation splits, respectively. 
The nine annotators are graduate students with engineering or linguistics background. 
These annotators were paid 10 Euro per hour for their participation.
The four expert annotators are doctoral and postdoctoral researchers of our group with a research focus on computer vision and multimodal analytics. 
Each annotator was shown a tweet text with its corresponding image and asked to answer the questions presented in Section~\ref{sec:guideline}. 
Exactly three annotators labeled each sample, and we used a majority vote to obtain the final label.

\subsubsection{Claim Categories}\label{subsubsec:dataset_labels} %Dataset Labels
We selected a total of \num{3400} tweets for manual annotation of training~(annotated by external annotators) and evaluation~(annotated by internal experts) splits. Each split contains an equal number of samples for the topics: \emph{COVID-19}, \emph{Climate Change}, and \emph{Technology}.
%Each annotator was shown a tweet text with its corresponding image and asked to answer the questions presented in Section~\ref{sec:guideline}. 
%Exactly three annotators labeled each sample, and we used a majority vote to obtain the final label.
% 
Labels for three types of claim\footnote{Here claim is a factually-verifiable claim not any claim} annotations are derived:
\begin{itemize}
	\setlength\itemsep{0em}
	% \item Binary claim classification: \textit{not a claim}, and \textit{claim}
	% \item Tertiary claim classification: \textit{not a claim}, \textit{claim but not check-worthy}, and \textit{check-worthy claim}
	% \item Visual claim classification: \textit{not a claim}, \textit{visually-irrelevant claim}, and \textit{visually-relevant claim}
	\item binary claim classes: \textit{not a claim}, and \textit{claim}
	\item tertiary claim classes: \textit{not a claim}, \textit{claim but not check-worthy}, and \textit{check-worthy claim}
	\item visual claim classes: \textit{not a claim}, \textit{visually-irrelevant claim}, and \textit{visually-relevant claim}
\end{itemize}

% We also observed that some tweets had barely any text (like one word) and no images, which are removed from the final dataset during annotation.

\subsubsection{Annotator Training}
The annotators were trained with detailed annotation guidelines, which included the definitions given in Section~\ref{sec:guideline} and multiple examples. To ensure the quality, we performed two dry runs using a set of samples (30-40) to annotate. Afterwards, the annotations were discussed to check agreements among annotators and the guidelines were refined based on the feedback.

\subsubsection{Inter-Annotator Agreement}

We measured the agreements between two groups of annotators using \textit{Krippendorff's alpha}~\cite{krippendorff2011computing}. The agreements were computed for the three types of annotations described in the previous section. For the training dataset group, we observe \num{0.53}, \num{0.39}, and \num{0.42} as agreement scores for the \textit{binary}, \textit{tertiary}, and \textit{visual claims}, respectively. 
For the test dataset group, we observe the following agreement scores: \num{0.57}, \num{0.47}, and \num{0.52} for three classifications, respectively.
The moderate agreement scores suggest that the problem of identifying check-worthy claims is partially a subjective task for both non-experts and experts.
%
% We observed conflicts for Q2 and Q3 when answer to Q1 by one annotator was 'no' and 'yes' by the others. 
% In this case, we resolve the conflict by giving priority to the claim but not check-worthy label in tertiary and visually-relevant claim label for visual claim classification. 
% This gives us two sets of splits, "with conflicts" and "without conflicts", for which we performed an experimental comparison (see Section~\ref{sec:experiments}).
%

While a majority is always possible for the binary claim classification that allows us to derive unambiguous labels, entirely different labels could be chosen for the tertiary and visually-relevant claim classification task since the annotators assign three possible classes.
Consequently, it is not  possible to derive a label with majority voting when each annotator selects a different option. 
In such cases, we resolve the conflict by prioritizing the \textit{claim but not check-worthy} class since check-worthiness is a stricter constraint and chosen by only one annotator, while two annotators agreed it is a claim. 
For visual claims, we select a \textit{visually-relevant claim} since it is possible that image and text are related, even when one annotator marked "no" to the claim question. 
A table and detailed explanation of the conflict cases is described in Appendix~\ref{app:conflict_resolution}.

%%% OLD PARTS %%%%
% We computed \textit{Krippendorff's alpha} as Inter-Annotator Agreement (IAA) separately for two groups of annotators. For the training dataset group, we observe 0.530, 0.389 and 0.419 as agreement scores for the three types of classification, respectively. For the testing dataset group, we observe 0.477, 0.373 and 0.439 as agreement scores. To improve the quality of the testing dataset, we remove conflicts for Q2 and Q3, which improves the agreement scores to 0.566, 0.469 and 0.520 for three classifications, respectively.The moderate agreement scores suggest that the problem of identifying claims and check-worthiness is a highly subjective task for both non-experts and experts.

\subsection{The MM-Claims Dataset}\label{subsec:mmclaim}
As a result of the annotation process, the \textit{Multimodal Claims (MM-Claims)} dataset\footnote{Source code is available at: \url{https://github.com/TIBHannover/MM_Claims} \\ Dataset (Tweet IDs) and labels are available at: \url{https://data.uni-hannover.de/dataset/mm_claims} \\ For complete labeled data access (Images and Tweets), please contact at \textit{gullal.cheema@tib.eu} or \textit{gullalcheema@gmail.com}} consists of \num{2815}~($T_C$ (training)) and \num{585}~($E_C$ (evaluation)) samples ($C$ in the subscript stands for "with resolved conflicts"). 
However, as discussed above, there are conflicting examples for the tertiary and visual claim labels. To train and evaluate our models on unambiguous examples, we derive a subset of \textit{Multimodal Claim (MM-Claims)} dataset that contains \num{2555}~($T$) and \num{525}~($E$) samples "without conflicts" where a majority vote can be taken. We divided the training set ($T_C$, $T$) in each case further into training and validation in a 90:10 split for hyper-parameter tuning. 
% Figure~\ref{fig:data_stats} shows the topic and class distributions in the labeled dataset.   

We noticed that one-third of the images in the dataset contains a considerable amount of overlaid text~(five or more words). 	As suggested by previous work ~\cite{DBLP:conf/www/CheemaHME21,DBLP:journals/corr/abs-2103-06304,DBLP:journals/corr/abs-2107-04313}, overlaid text in images should be considered in addition to tweet text and other image content.
Specifically, the images with overlaid text not only act as related information to the tweet text but are sometimes the central message of the tweet.
We used Tesseract-OCR~\cite{tesserocr} to select images that contain five or more words in their overlay text. In an internal pre-test with 100 images, we observed that Tesseract-OCR produced more random (and incorrect) text from images than Google Vision API. 
To reduce the incorrect text, we ran Google Vision API on the selected images (avoiding unnecessary costs) in the second step that resulted in a better quality OCR detected text.
Besides the labeled dataset, we will also provide the images, tweet text, and the overlay text~(extracted using OCR methods as described above) of the unlabeled portion of the dataset.
\section{Experimental Setup and Evaluation}\label{sec:experiments}
In this section, we describe the features, baseline models, and the comprehensive experiments using our novel dataset. 
We test a variety of features and recent multimodal state-of-the-art models. % for our use case.

\subsection{Features}
% Here, we give details of state-of-the-art image, text and multimodal features that are used to encode image-text pairs from our dataset. 

\noindent\textbf{Pre-processing:} For images, we use the standard pre-processing of resizing and normalizing an image, whereas text is cleaned and normalized according to \citet{cheema2020checksquare} using the  Ekphrasis~\cite{DBLP:conf/semeval/BaziotisPD17a} tool. 
Besides digits and alphabets, we also keep punctuation to reflect the syntax and style of a written claim.

\noindent\textbf{Image Features:}
For image encoding, we use a \emph{ResNet-152}~\cite{he2016deep} model trained  %object recognition 
on \emph{ImageNet}~\cite{russakovsky2015imagenet} and extract the 2048-dimensional feature vector from the last pooling layer. 
%Such pre-trained model features are widely considered a good image representation for several unimodal and multimodal tasks.

\noindent\textbf{Text Features}:
For encoding tweet and OCR text, we test \textit{BERT}~\cite{devlin2018bert} uncased models to extract contextual word embeddings. 
% 
%For SVM~\cite{suykens1999least} (Support Vector Machine) experiments, we employ a pooling strategy \cite{cheema2020checksquare,DBLP:conf/www/CheemaHME21}, where last four layers' outputs are added to get the word embeddings. 
For classification using Support Vector Machine (SVM, ~\cite{DBLP:journals/ml/CortesV95}), we employ a pooling strategy by adding the last four layers' outputs and then average them to obtain the final 768-dimensional vector.
%The word embeddings are then averaged to obtain the final 768-dimensional vector for the tweet text.

\noindent\textbf{Multimodal Features}:
We use the following two pre-trained image-text representation learning architectures to extract multimodal features.

\noindent The \textbf{\emph{ALBEF}} (ALign BEfore Fuse) embedding ~\cite{DBLP:journals/corr/abs-2107-07651} results from a recent  multimodal state-of-the-art model for vision-language downstream tasks. 
It is trained on a combination of several image captioning datasets ($\sim$\num{14} million image-text pairs) and uses \textit{BERT} and a visual transformer~\cite{DBLP:conf/iclr/DosovitskiyB0WZ21} for text and image encoding, respectively.
It produces a multimodal embedding of 768 dimensions.

\noindent The \textbf{\emph{CLIP}} (Contrastive Language-Image Pretraining) model ~\cite{DBLP:conf/icml/RadfordKHRGASAM21}
is trained without any supervision on \num{400} million image-text pairs.
% Thus, the images in the data are not restricted and contain words or sentences. 
We evaluate several image encoder backbones including \emph{ResNet} and vision transformer~\cite{DBLP:conf/iclr/DosovitskiyB0WZ21}. The \emph{CLIP} model outputs two embeddings of same size, i.e., the image ($CLIP_I$) and the text ($CLIP_T$) embedding, while $CLIP_{I \oplus T}$ denotes the concatenation of two embeddings.

\subsection{Training Baselines}
In the following, we describe training details, hyper-parameters, input combinations, and different baseline models' details. % We produce all the experiments using PyTorch library.

\subsubsection{SVM}
To obtain unimodal and multimodal embeddings for our experiments, 
we first use PCA (Principal Component Analysis) to reduce the feature size and train a SVM model with the \textit{RBF} kernel. 
We perform grid search over PCA energy (\%) conservation, regularization parameter \textit{C} and \textit{RBF} kernel’s \textit{gamma}. 
The parameter range for PCA varies from 100\% (original features) to 95\% with decrements of 1. 
The parameter range for \textit{C} and \textit{gamma} vary between \num{-1} to \num{1} on a log-scale with 15 steps. 
For multimodal experiments, image and text embeddings are concatenated before passing them to PCA and SVM. 
We normalize the final embedding so that \textit{l2} norm of the vector is 1.

\subsubsection{BERT and ALBEF Fine-tuning (FT)}
We experiment with fine-tuning the last few layers of unimodal and multimodal transformer models to get a strong multimodal baseline and see whether introducing cross-modal interactions improves claim detection performance. We fine-tune the last layers of both the models and report the best ones in Table~\ref{tab:claimdet_1}. Additional experimental results on fine-tuned layers are provided in Appendix~\ref{app:exp_num_layers}. For fine-tuning, we limit the tweet text to the maximum number of tokens (91) seen in a tweet in the training data and pad the shorter tweets with zeros. Hyper-parameter details for fine-tuning are provided in the Appendix~\ref{sec:hyp_ft}.

\subsubsection{Models with OCR Text}
To incorporate OCR text embeddings into our models, we experiment with two strategies for embedding generation and one strategy to fine-tune models. To obtain an embedding for SVM models, we experimented with concatenating the OCR embedding to image and tweet text embeddings as well as adding the OCR embedding directly to tweet text embedding. 
To fine-tune the models, we concatenate the OCR text to tweet text and limit the OCR text to 128 tokens.

\subsubsection{State-of-the-Art Baselines}
We compare our models with two state-of-the-art approaches for multimodal fake news detection. 

\noindent\textbf{\emph{MVAE}}~\cite{khattar2019mvae} is a multimodal variational auto-encoder model that uses a multi-task loss to minimize the reconstruction error of individual modalities and task-specific cross-entropy loss for classification. 
We use the publicly available source code and hyper-parameters for our task.

\noindent\textbf{\emph{SpotFake}}~\cite{DBLP:conf/bigmm/SinghalS0KS19} is a model built as a shallow multimodal neural network on top of \emph{VGG-19} image and \emph{BERT} text embeddings using a cross-entropy loss. We re-implement the model in PyTorch and use the hyper-parameter settings given in the paper.

\subsection{Results}
We report accuracy (Acc) and Macro-F1 (F1) for binary (BCD) and tertiary claim detection (TCD) in Table~\ref{tab:claimdet_1}. 
We also present the fraction (in \%) of visually-relevant and visually-irrelevant (textual only) claims retrieved by each model in Table~\ref{tab:vistext_1}. Please note that in Table~\ref{tab:claimdet_1} and Table~\ref{tab:claimdet_2}, BCD results are shown for only one split ($T_C \rightarrow E_C$), because there are no conflicts in the labels for binary claim classification. Although we do not train the models specifically to detect visual claim labels, we analyze the fraction of retrieved samples in order to evaluate the bias of binary classification models towards a modality.

% \bgroup
\setlength{\tabcolsep}{4pt} % Default value: 6pt
\begin{table}[t]
	\footnotesize
	% \begin{tabular}{|l|*{4}{p{15mm}|}|*{4}{p{15mm}|}}
		\begin{tabularx}{\linewidth}{|X|*{2}{>{\centering\arraybackslash}p{5mm}|}|*{4}{>{\centering\arraybackslash}p{5mm}|}}
			% \begin{tabularx}{\linewidth}{|X|*{6}{>{\centering\arraybackslash}c|}|*{6}{>{\centering\arraybackslash}c|}}
				\hline
				\textbf{Task} $\rightarrow$         & \multicolumn{2}{c||}{\textbf{BCD}} & \multicolumn{4}{c|}{\textbf{TCD}} \\ \hline
				\textbf{Data Splits} $\rightarrow$  & \multicolumn{2}{c||}{\textbf{$T_C \rightarrow E_C$}}  & \multicolumn{2}{c|}{\textbf{$T \rightarrow E_{C}$}} & \multicolumn{2}{c|}{\textbf{$T_{C} \rightarrow E_{C}$}} \\ \hline
				\textbf{Models} $\downarrow$                            & \textbf{Acc}  & \textbf{F1} & \textbf{Acc}  & \textbf{F1} & \textbf{Acc}  & \textbf{F1}   \\ \hline
				Random                                                  & 50.7         & 50.2       & 33.3         & 30.6       & 33.3         & 30.6         \\ \hline
				Majority                                                & 62.7         & 38.5       & 56.2         & 35.9       & 56.2         & 35.9        \\ \hline
				\hline
				ImageNet                                                & 63.1         & 62.6       & 58.3         & 42.9       & 58.5         & 43.9         \\ \hline
				CLIP$_I$                                                & 70.0         & 69.8       & 64.1         & 50.5       & 62.4         & 48.7         \\ \hline
				BERT                                                    & 80.5         & 79.9      & 71.9         & 54.1       & 69.6         & 59.8         \\ \cline{2-7}
				$\hookrightarrow$ FT                                    & 80.9         & 80.1      & 72.5         & 54.5       & \textbf{75.4}         & \textbf{64.6}         \\ \hline
				CLIP$_T$                                                & 75.6         & 74.7       & 70.6         & 53.4       & 67.4         & 54.5         \\ \hline
				\hline
				BERT $\oplus$ ImageNet                                  & \textbf{81.4}         & 80.9       & 72.7         & 57.6       & 71.6         & 56.9         \\ \cline{2-7}
				$\hookrightarrow$ $\oplus$ OCR                          & 80.9         & 80.4       & 72.8         & 58.2       & 71.9         & 58.6         \\ \hline
				CLIP$_{I \oplus T}$                                     & 77.8         & 77.4       & 71.6         & 52.9       & 68.4         & 54.6         \\ \hline
				CLIP$_{I}$ $\oplus$ BERT                                & 80.3         & 79.7       & 72.7         & 57.9       & 69.4         & 59.7         \\ \hline
				ALBEF                                                   & 76.9         & 76.5       & 71.5         & 56.1       & 65.6         & 57.3         \\ \cline{2-7}
				$\hookrightarrow$ FT                                    & 80.2         & 79.7       & \textbf{74.5}         & \textbf{60.7}       & 72.5        & 61.0         \\ \cline{2-7}
				$\hookrightarrow \oplus$ OCR $\oplus$ FT                & \textbf{81.4}         & \textbf{81.1}       &     72.7     &    58.2    & 73.0         & 60.8         \\ \hline
				\hline
				MVAE                                                    & 64.1      & 62.9       & 60.0         & 41.2       & 59.7         & 44.8         \\ \hline
				SpotFake                                                & 71.8         & 71.4       & 67.0         & 49.5       & 66.3         & 52.2         \\ \hline
			\end{tabularx}
			\caption{Accuracy~(Acc) and Macro-F1~(F1) for binary~(BCD) and tertiary claim detection~(TCD) in percent~[\%]. As described in Section~\ref{subsec:mmclaim}, we use the training split~($T$) with resolved~(index $C$) and without~(no index) conflicts, and evaluation~(test) split~($E_C$) with conflicts. This evaluation split reflects the real-world scenario for the subjective task of tertiary claim classification~(TCD). Unless FT (fine-tuning) is written, all models (except MVAE and SpotFake) are SVM models trained on extracted features.}\label{tab:claimdet_1}
		\end{table}
		% \egroup
% \bgroup
\setlength{\tabcolsep}{2pt} % Default value: 6pt
\begin{table}[!b]
	\footnotesize
	%results for True Positive Rate~(TPR) are reported in percent~[\%].}
% \begin{tabular}{|l|*{4}{p{15mm}|}|*{4}{p{15mm}|}}
	\begin{tabularx}{\linewidth}{|X|*{4}{>{\centering\arraybackslash}p{10mm}|}}
		\hline
		\textbf{Data Splits} $\rightarrow$    & \multicolumn{2}{c|}{\textbf{$T \rightarrow E_{C}$}} & \multicolumn{2}{c|}{\textbf{$T_{C} \rightarrow E_{C}$}} \\ \hline
		% \textbf{Models} $\downarrow$    & \textbf{Visual (111)}                 & \textbf{Textual (145)}        & \textbf{Visual (76)}                 & \textbf{Textual (120)}   & \textbf{Visual (111)}                 & \textbf{Textual (145)}  & \textbf{Visual (76)}                 & \textbf{Textual (120)}               \\
		\textbf{Models} $\downarrow$                            & \textbf{V (111)}  & \textbf{T (145)}      & \textbf{V (111)}   & \textbf{T (145)}  \\
		\hline
		ImageNet                                              & 35.1             & 39.3                 & 61.3             & 57.9             \\ \hline
		CLIP$_I$    & 70.3             & 67.6                 & \textbf{76.6}             & 73.8             \\ \hline
		BERT                                           & 49.6             & 76.6                 & 57.7             & 82.1             \\ \cline{2-5}
		$\hookrightarrow$ FT                                    & 52.3             & 75.9                 & 55.9             & \textbf{82.8} \\ \hline
		CLIP$_T$                                                & 46.9             & 73.1                 & 54.9             & 73.1             \\ \hline
		\hline
		BERT $\oplus$ ImageNet                           & 57.7             & 66.2                 & 71.2             & 77.9             \\ \cline{2-5}
		$\hookrightarrow$ $\oplus$ OCR                          & 65.8             & 75.9                 & 71.2             & 79.3             \\ \hline
		CLIP$_{I \oplus T}$                                     & 65.8             & 66.9                 & 72.9             & 75.2             \\ \hline
		CLIP$_{I} \oplus$ BERT                           & 57.7             & 72.4    & 57.7     & \textbf{82.8}             \\ \hline
		ALBEF   & 61.2             & 75.2                 & 63.9             & 77.9             \\ \cline{2-5}
		$\hookrightarrow$ FT                                    & 62.2             & 77.2                 &       70.3      &       78.6       \\ \cline{2-5}
		$\hookrightarrow \oplus$ OCR $\oplus$ FT                & \textbf{71.2}  &   \textbf{79.3}    & 75.7    & 82.1 \\ \hline
	\end{tabularx}
	\caption{Visually-relevant ~(V) and visually-irrelevant~(text-only)~(T) claim detection evaluation. The number of test samples is reported in brackets and the fraction, how many of them were retrieved, is given in percent~[\%]. The underlying models are trained for binary claim detection (BCD). The labels for visual relevance are only used for retrieval evaluation.}\label{tab:vistext_1}
\end{table}
% \egroup

\subsubsection{Impact of Annotation Disagreements}\label{sec:agreement}
As mentioned in Section~\ref{sec:dataset}, we observed disagreements in the annotated data that reflect the real-world difficulty and subjectivity of the problem. 
Therefore, we analyze the effect of keeping ($T_C$, $E_C$) and removing ($T$, $E$) conflicting examples in training and evaluation data splits (Table~\ref{tab:claimdet_1},~\ref{tab:claimdet_2}). 
The findings are as follows: 1) multimodal models are more sensitive to the conflict resolution strategy as most have lower accuracy when trained on $T_C$ but relatively better F1 score. 
On the contrary, visual and textual models perform better on both metrics with training on $T_C$, 2) overall, training on $T_C$ with conflict resolution is a better strategy with a higher F1 score, i.e., better on claim and check-worthiness (fewer samples) detection; and 3) when comparing all the cross-split experiments in Table~\ref{tab:claimdet_1} and Table~\ref{tab:claimdet_2}, multimodal models perform the best in case of "without conflicts" $T$ and $E$ splits. 
The latter two observations also apply to retrieval of visually-relevant and visually-irrelevant claims in Table~\ref{tab:vistext_1} and Table~\ref{tab:vistext_2}.

\begin{figure*}[th!]
	\centering
	\includegraphics[width=\linewidth]{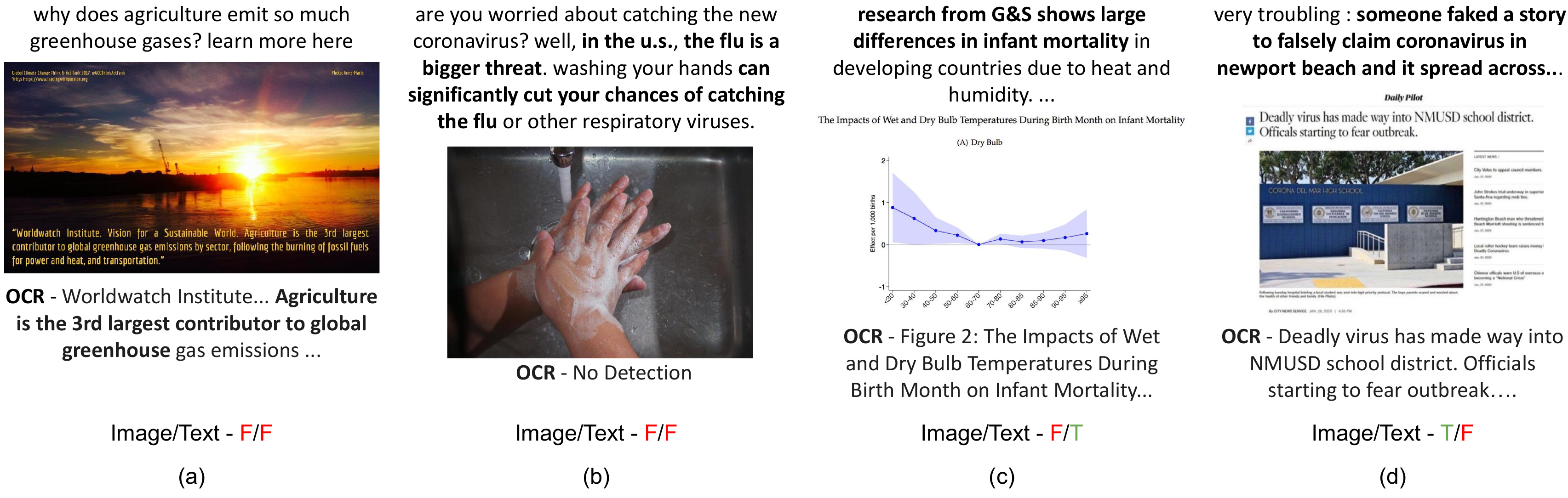}
	\caption{Qualitative examples where our best multimodal model classifies correctly and unimodal models do not. \textit{F} - false classification, \textit{T} - true classification. }
	\label{fig:qualit}
\end{figure*}

\subsubsection{Results for Unimodal Models}
For image-based models, \emph{CLIP$_I$} performs (\num{70.0}, \num{69.8}) considerably better than \emph{ResNet-152}'s \emph{ImageNet} (\num{63.1}, \num{62.6})  features in terms of both accuracy and F1 metrics (Table~\ref{tab:claimdet_1}, block 2). This result is compliant to previous work~\cite{DBLP:journals/corr/abs-2107-04313} where the task has a variety of information and text in images. 
It is further exaggerated and clearly observable in Table~\ref{tab:vistext_1} where fraction of visually-relevant claims retrieved using \emph{CLIP$_I$} (\num{70.3}) is higher and comparable to fine-tuned \emph{ALBEF} $\oplus$ OCR (\num{71.2}).

For text-based models, fine-tuning (FT) \emph{BERT} gives the best performance, better than any other unimodal model. This result indicates that the problem is inherently a text-dominant task. The model also retrieves the most visually-irrelevant claims when trained on $T_C$. It should be noted that textual models can still identify visually-relevant claims since they can have a claim or certain cues in the tweet text that refer to the image. Finally, the \emph{CLIP$_T$} features perform considerably worse than \emph{BERT} features, possibly because \emph{CLIP} is limited to short text (75 tokens) and is not trained like vanilla \emph{BERT} on a large text corpus.

\subsubsection{Results for Multimodal Models}
For multimodal models, %the results show the promise of using multiple modalities for claim detection. 
the combination of \emph{BERT} and \emph{ResNet-152} features performs slightly better ($0.5-1\%$) on two metrics in Table~\ref{tab:claimdet_1} on full dataset in binary task and with $T$ split training in case of tertiary. 
Although this gain is not impressive, the benefit of combining two modalities is more obvious in identifying visually-relevant claims ($>10\%$) in Table~\ref{tab:vistext_1}, which comes at the cost of a lower fraction of visually-irrelevant claims.
Similarly with \emph{CLIP}, the combination of image and text features (CLIP$_{I \oplus T}$) improves the overall accuracy from CLIP$_I$ or CLIP$_T$. 
However, we do not see the same result for identifying visually-relevant claims ($<4-5\%$). 
We also experiment with the combination of \emph{BERT} features with \emph{CLIP}'s image features, which improves the overall accuracy further but indicates that the model relies strongly on text (65.8 vs. 57.7 visual retrieval \%) rather than the combination. The stronger reliance on text is possibly not a trait of the model alone, but could be also caused by an incompatibility of \emph{BERT} and CLIP$_I$ features. %, instead of CLIP$_T$.

Finally, we achieve the best performance (by $1-4\%$) on binary and tertiary (when trained on $T$) claim detection by fine-tuning the \emph{ALBEF} with and without OCR, respectively (Table~\ref{tab:claimdet_1}, block 3, last row). 
While the benefit of using OCR text in SVM models is not optimal and not considerably helpful, OCR addition to \emph{ALBEF} retrieves the maximum number of visually-relevant claims ($71.2\%$) without losing much on visually-irrelevant claims ($79.3\%$) when trained on $T$ (Table~\ref{tab:vistext_1}, block 2, last row).
These results point towards a major challenge of combining multiple modalities and retaining intra-modal information (and influence) for the task at hand. As noted in section~\ref{sec:agreement}, an interesting result is that \emph{ALBEF} in particular is less robust to resolved conflicts (split $T_C$) in the data when compared to just using \emph{BERT}. On closer inspection, these conflicts are mostly caused by the image relevance to the text. The gap is further exaggerated in Table~\ref{tab:claimdet_2}, where \emph{ALBEF} performs much better than \emph{BERT}, when conflict examples are removed from both training and evaluation.
Figure~\ref{fig:qualit} shows a few examples where our best multimodal model correctly classifies, whereas unimodal models based on either image or text do not. 
All the samples in the figure have images that have some connection to the tweet text. 
The image in Figure~\ref{fig:qualit}b has a connection to one of the words or phrases (e.g., washing your hands) in the tweet text but is not relevant for the claim itself. 
Figure~\ref{fig:qualit}a includes an image with the claim itself and a very generic scene in the background. Both image and text in Figure~\ref{fig:qualit}c and Figure~\ref{fig:qualit}d are relevant, and the image acts as evidence and additional information. 
In all these examples, a rich set of information extraction and complex cross-modal learning is required to identify claims in multimodal tweets. When comparing results of recent state-of-the-art architectures for fake news detection, SpotFake~\cite{DBLP:conf/bigmm/SinghalS0KS19} does considerably better than MVAE~\cite{khattar2019mvae} but worse than any of our baseline models.
% While both the models have different loss functions, primary reason for lower performance in SpotFake seems to be simple concatenation of features before classification. Whereas, both models have large number of parameters that cannot be efficiently trained from scratch on a smaller dataset.

\section{Conclusions}\label{sec:conclusion}
In this paper, we have presented a novel \textit{MM-Claims} dataset to foster research on multimodal claim analysis. The dataset has been curated from Twitter data and contains more than \num{3000} manually annotated tweets for three tasks related to claim detection across three topics, \emph{COVID-19}, \emph{Climate Change}, and \emph{Technology}. 
We have evaluated several baseline approaches and compared them against two state-of-the-art fake news detection approaches. 
Our experimental results suggest that the fine-tuning of pre-trained multimodal and unimodal architectures such as \textit{ALBEF} and \textit{BERT} yield the best performance. 
We also observed that the overlaid text in images is important in information dissemination, particularly for claim detection. To this end, we evaluated a couple of strategies to incorporate OCR text into our models, which yielded a much better trade-off between identifying visually-relevant and visually-irrelevant (text-only) claims. 

In the future, we will explore other and novel architectures for multimodal representation learning and other information extraction techniques to incorporate individual modalities better. 
%Specifically, having a better image representation encodes images with text, graphs, and other object and scene information. 
We also plan to investigate fine-grained overlaps of concepts and meaning in image and text, %as pointed out in the examples in Figure~\ref{fig:qualit} 
and expand the dataset to COVID-19 related sub-topics and specific climate change events.
%Lastly, expanding the dataset to COVID-19 related sub-topics and specific climate change events is also a subject for future work.
% and learn a common cross-modal representation.

\section*{Acknowledgements}
This work was funded by European Union’s Horizon 2020 research and innovation programme under the Marie Skłodowska-Curie grant agreement no 812997 (CLEOPATRA project), and by the German
Federal Ministry of Education and Research (BMBF, FakeNarratives project, no. 16KIS1517). 

% Entries for the entire Anthology, followed by custom entries
\bibliography{anthology,references}
\bibliographystyle{acl_natbib}

\appendix

\newpage

\section{Appendix}\label{sec:appendix}
In the following we include additional hyper-parameter details~(\ref{sec:hyp_ft}) and experimental results~(\ref{sec:add_exps}), additional dataset and annotation process details~(\ref{sec:add_data}), and some annotated tweets for multimodal claim detection~(\ref{sec:add_samples}).

\subsection{Hyper-parameters and other details}\label{sec:hyp_ft}
For fine-tuning \emph{BERT} and \emph{ALBEF}, we use a batch-size of 16 and 8 (size constraints) respectively. We train the models for five epochs and use the best (on validation set) performing (accuracy) model for evaluation. For \emph{BERT}, a dropout with the ratio of \num{0.2} is applied before the classification head. Further, we use AdamW~\cite{DBLP:conf/iclr/LoshchilovH19} as the optimizer with a learning rate of $3e-5$ and a linear warmup schedule. The learning rate is first linearly increased from $0$ to $3e-5$ for iterations in the first epoch and then linearly decreased to $0$ for the rest of the iterations in $4$ epochs. For \emph{ALBEF}, we use the recommended fine-tuning hyper-parameters and settings from the publicly available code.

\subsection{Additional Experimental Results}\label{sec:add_exps}

\subsubsection{CLIP Variants}
We experiment with \emph{CLIP}'s three variants that use different visual encoder backbones, ResNet-50 (RN50), ResNet-50x4 (RN504) and a vision transformer (ViT-B/16)~\cite{DBLP:conf/iclr/DosovitskiyB0WZ21} with \emph{BERT} as textual encoder backbone. We select the models for textual and multimodal SVM experiments based on the performance (higher accuracy) using features from the visual encoders. Table~\ref{tab:clip} shows different visual encoders' features (with SVM) performance on binary and tertiary claim detection.

It should be noted that just like \emph{ALBEF}, \emph{CLIP} models can be fine-tuned with image-text tweet pairs for binary and tertiary tasks. However, when we experimented with fine-tuning the last few layers of \emph{CLIP} with a classification head on top, it always performed worse than using extracted features for classification with SVM. This phenomenon is probably because of our relatively smaller sized labeled dataset, which is not enough for fine-tuning \emph{CLIP} for the task. 

\begin{table}[t]
	\footnotesize
	% \begin{tabular}{|l|*{4}{p{15mm}|}|*{4}{p{15mm}|}}
		\begin{tabularx}{\linewidth}{|X|*{2}{>{\centering\arraybackslash}p{5mm}|}|*{4}{>{\centering\arraybackslash}p{5mm}|}}
			% \begin{tabularx}{\linewidth}{|X|*{6}{>{\centering\arraybackslash}c|}|*{6}{>{\centering\arraybackslash}c|}}
				\hline
				\textbf{Task} $\rightarrow$         & \multicolumn{2}{c||}{\textbf{BCD}} & \multicolumn{4}{c|}{\textbf{TCD}} \\ \hline
				\textbf{Data Splits} $\rightarrow$  & \multicolumn{2}{c||}{\textbf{$T_C \rightarrow E_C$}}  & \multicolumn{2}{c|}{\textbf{$T \rightarrow E_{C}$}} & \multicolumn{2}{c|}{\textbf{$T_{C} \rightarrow E_{C}$}} \\ \hline
				\textbf{Models} $\downarrow$                            & \textbf{Acc}  & \textbf{F1} & \textbf{Acc}  & \textbf{F1} & \textbf{Acc}  & \textbf{F1}   \\ \hline
				\hline
				RN50                                                & 66.3         & 65.7       & 64.1         & 50.6       & \textbf{62.4}         & \textbf{48.7}         \\ \hline
				RN50x4                                               & \textbf{70.0}         & \textbf{69.9}       & 61.5         & \textbf{51.5}       & 61.4         & 48.5         \\ \hline
				ViT-B/16                                                  & 68.6         & 68.4      & \textbf{64.3}         & 49.8       & 59.7         & 48.3         \\    \hline
			\end{tabularx}
			\caption{CLIP's different visual encoder backbones features' performance evaluation. Accuracy~(Acc) and Macro-F1~(F1) for binary~(BCD) and tertiary claim detection~(TCD) in percent~[\%]. As described in Section~\ref{subsec:mmclaim}, we use the Training Split~($T$) and Evaluation~(Testing) Split~($E$) with resolved~(index $C$) and without~(no index) conflicts.}\label{tab:clip}
		\end{table}
		% \egroup

\subsubsection{Results for "without conflicts" ($E$) Evaluation Split}
In Section~\ref{sec:experiments}, we show results for tertiary claim detection (TCD) on evaluation splits "with resolved conflicts" ($E_C$) by training on $T$ and $T_C$. Here in Table~\ref{tab:claimdet_2}, we show the evaluation on "without conflicts" evaluation split ($E$). As with evaluation on $E_C$, multimodal models are more sensitive to training on $T_C$ where conflict resolution strategy causes the accuracy to drop for all models. However, \emph{CLIP} and \emph{ALBEF} models, in this case, have higher F1-score (as well as accuracy) when trained on $T$. Even with less training data, the models perform better and best among all evaluated multimodal models. In the case of training on $T_C$, \emph{BERT} performs the best, which is closely followed by \emph{ALBEF} with OCR text.

As described in section~\ref{sec:agreement}, the evaluation of retrieved visually-relevant and visually-irrelevant claims on $E$ follows the evaluation on $E_C$. Even though $CLIP_I$ and fine-tuned \emph{BERT} retrieves the most amount of two types of claims, all models do better when trained on $T_C$ than on $T$.

Overall, for a realistic scenario, training on $T_C$ gives the best performance trade-off between Acc, F1 and retrieved claims for multimodal models.

% \bgroup
\setlength{\tabcolsep}{4pt} % Default value: 6pt
\begin{table}[t]
	\footnotesize
	% \begin{tabular}{|l|*{4}{p{15mm}|}|*{4}{p{15mm}|}}
		\begin{tabularx}{\linewidth}{|X|*{4}{>{\centering\arraybackslash}p{10mm}|}}
			% \begin{tabularx}{\linewidth}{|X|*{6}{>{\centering\arraybackslash}c|}|*{6}{>{\centering\arraybackslash}c|}}
				\hline
				\textbf{Task} $\rightarrow$         & \multicolumn{4}{c|}{\textbf{TCD}} \\ \hline
				\textbf{Data Splits} $\rightarrow$  & \multicolumn{2}{c|}{\textbf{$T \rightarrow E$}} & \multicolumn{2}{c|}{\textbf{$T_{C} \rightarrow E$}} \\ \hline
				\textbf{Models} $\downarrow$                             & \textbf{Acc}  & \textbf{F1} & \textbf{Acc}  & \textbf{F1}   \\ \hline
				Random                                                  & 33.7         & 28.2       & 33.7         & 28.2         \\ \hline
				Majority                                                &  62.7         & 38.5       & 62.7         & 38.5        \\ \hline
				\hline
				ImageNet                                                &  62.5         & 40.9       & 62.5         & 42.1         \\ \hline
				CLIP$_I$                                                &  68.9         & 50.2       & 67.2         & 48.7         \\ \hline
				BERT                                                    &  77.9         & 52.9       & 72.8         & 56.9         \\ \cline{2-5}
				$\hookrightarrow$ FT                                    &  78.3         & 51.2       & \textbf{79.2}         & \textbf{61.4}         \\ \hline
				CLIP$_T$                                                &  77.3         & 54.4       & 71.6         & 52.3         \\ \hline
				\hline
				BERT $\oplus$ ImageNet                                  &  77.5         & 56.0       & 77.0         & 56.9         \\ \cline{2-5}
				$\hookrightarrow$ $\oplus$ OCR                          &  77.7         & 55.0       & 76.6         & 55.8         \\ \hline
				CLIP$_{I \oplus T}$                                     &  77.5         & 56.4       & 73.0         & 52.6         \\ \hline
				CLIP$_{I}$ $\oplus$ BERT                                &  77.9         & 53.3       & 72.6         & 56.8         \\ \hline
				ALBEF                                                   &  76.6         & 55.0       & 67.6         & 52.7         \\ \cline{2-5}
				$\hookrightarrow$ FT                                    &  \textbf{80.0}         & 63.3       & 76.8         & 59.7         \\ \cline{2-5}
				$\hookrightarrow \oplus$ OCR $\oplus$ FT                &  78.7     &    \textbf{63.5}    & 77.5         & 59.9         \\ \hline
				\hline
				MVAE                                                    &  64.8         & 40.7       & 62.9         & 43.2         \\ \hline
				SpotFake                                                &  72.8         & 49.7       & 70.7         & 50.4         \\ \hline
			\end{tabularx}
			\caption{Accuracy~(Acc) and Macro-F1~(F1) for tertiary claim detection~(TCD) in percent~[\%]. As described in Section~\ref{subsec:mmclaim}, we use the Training Split~($T$) and Evaluation~(Testing) Split~($E$) with resolved~(index $C$) and without~(no index) conflicts. Additional results on evaluation split without conflicts ($E$). Unless FT (fine-tuning) is written, all models (except MVAE and SpotFake) are SVM models trained on extracted features.}\label{tab:claimdet_2}
		\end{table}
		% \egroup

% \bgroup
\setlength{\tabcolsep}{2pt} % Default value: 6pt
\begin{table}[!th]
	\footnotesize
	%results for True Positive Rate~(TPR) are reported in percent~[\%].}
% \begin{tabular}{|l|*{4}{p{15mm}|}|*{4}{p{15mm}|}}
	\begin{tabularx}{\linewidth}{|X|*{4}{>{\centering\arraybackslash}p{10mm}|}}
		\hline
		\textbf{Data Splits} $\rightarrow$    & \multicolumn{2}{c|}{\textbf{$T \rightarrow E$}} & \multicolumn{2}{c|}{\textbf{$T_{C} \rightarrow E$}} \\ \hline
		% \textbf{Models} $\downarrow$    & \textbf{Visual (111)}                 & \textbf{Textual (145)}        & \textbf{Visual (76)}                 & \textbf{Textual (120)}   & \textbf{Visual (111)}                 & \textbf{Textual (145)}  & \textbf{Visual (76)}                 & \textbf{Textual (120)}               \\
		\textbf{Models} $\downarrow$                            & \textbf{V (76)}  & \textbf{T (120)}      & \textbf{V (76)}   & \textbf{T (120)}  \\
		\hline
		ImageNet                                              & 39.8             & 39.2                 & 67.1             & 58.3             \\ \hline
		CLIP$_I$                                            & 72.4             & 69.2                 & \textbf{78.9}             & 76.7             \\ \hline
		BERT                                           & 52.6             & 80.0                 & 61.8             & 85.0             \\ \cline{2-5}
		$\hookrightarrow$ FT                                    & 53.9             & 79.2                & 57.9             & \textbf{85.8} \\ \hline
		CLIP$_T$                                                & 51.3             & 76.7                 & 60.5             & 76.7             \\ \hline
		\hline
		BERT $\oplus$ ImageNet                           & 63.2             & 68.3                 & 75.0             & 80.8             \\ \cline{2-5}
		$\hookrightarrow$ $\oplus$ OCR                          & 69.7             & 78.3                 & 75.0             & 81.7             \\ \hline
		CLIP$_{I \oplus T}$                                     & 68.4             & 70.0                 & 76.3             & 78.3             \\ \hline
		CLIP$_{I} \oplus BERT$                           & 60.5             & 75.0    & 60.5    & 85.0             \\ \hline
		ALBEF                                           & 63.2             & 77.5                 & 65.8             & 80.8             \\ \cline{2-5}
		$\hookrightarrow$ FT                                    & 65.8             & 79.2                 &       75.0      &       80.8       \\ \cline{2-5}
		$\hookrightarrow \oplus$ OCR $\oplus$ FT                & \textbf{76.3}  &   \textbf{82.5}    & 77.6    & 85.0 \\ \hline
	\end{tabularx}
	\caption{Visually-relevant ~(V) and visually-irrelevant~(T) claim detection evaluation. The amount of test samples is reported in brackets and the fraction, how many of them were retrieved, is given in percent~[\%]. Additional results on evaluation split without conflicts ($E$).  The underlying models are trained for binary claim detection (BCD). The labels for visual relevance are only used for retrieval evaluation.}\label{tab:vistext_2}
\end{table}
% \egroup

\subsubsection{Confusion Matrix}

\begin{figure}[h!]
\centering
    \centering
    \includegraphics[width=\linewidth]{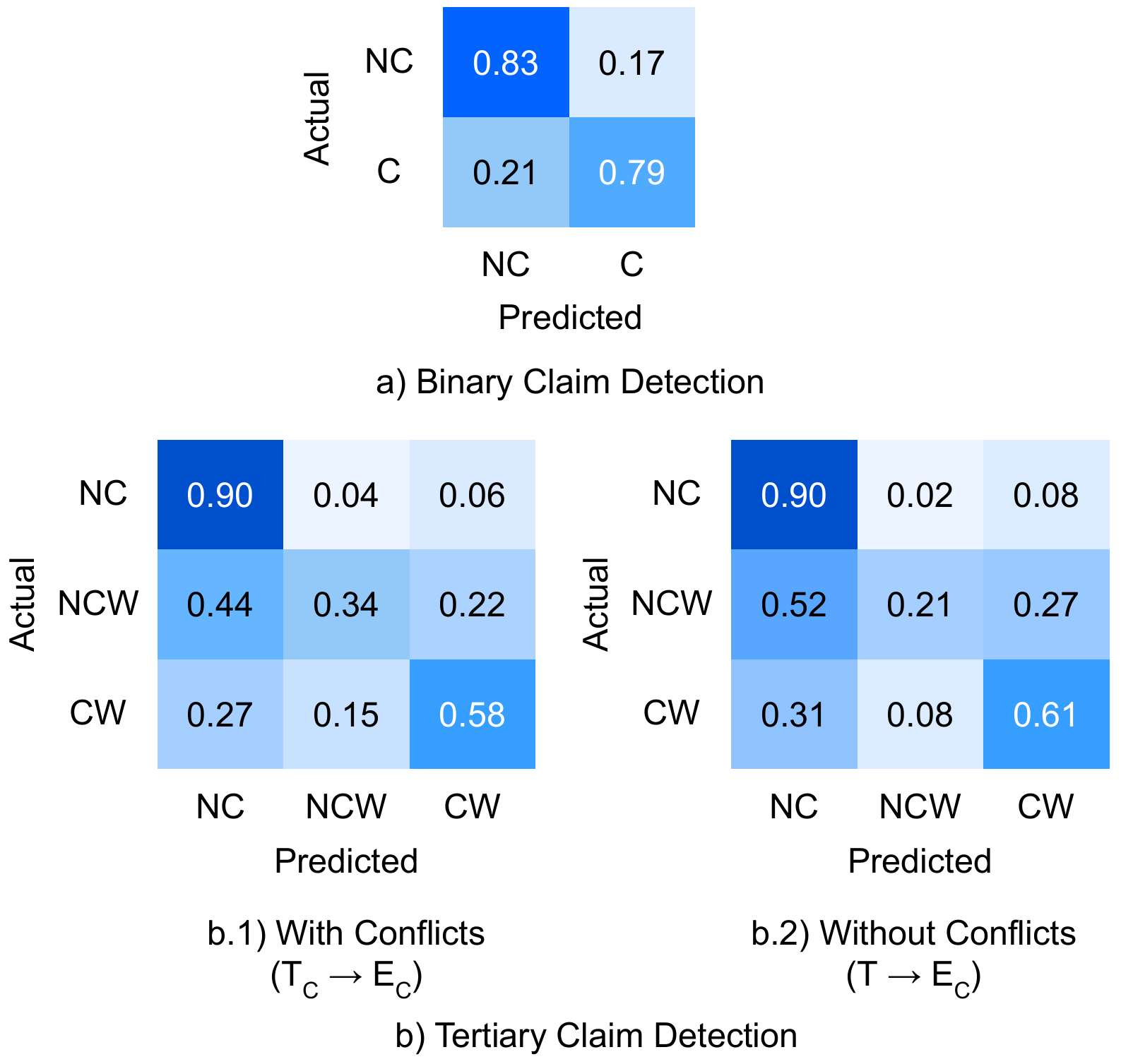}
    \caption{Normalized (by row) Confusion Matrices for the Binary and Tertiary Claim Classification Tasks. NC: Not-Claim, NCW: Not-check-worthy-Claim, C: Claim, CW: check-worthy-Claim}\label{fig:confmat}
\end{figure}

Following the results on $E_C$ in section~\ref{sec:experiments} for binary and tertiary tasks, we show normalized (by row) confusion matrices based on predictions from the $ALBEF\oplus OCR \oplus FT$ model. Figure~\ref{fig:confmat}a is the confusion matrix on $E_C$ for binary claim detection (BCD). Whereas, Figure~\ref{fig:confmat}b shows the matrices on $E_C$ with training on $T_C$ (b.1) and $T$ (b.2). Although the not-claim's true positives remain the same, confusion for the not-check-worthy and check-worthy class is less severe when trained on $T_C$.

\subsubsection{Ablation on OCR length}
\begin{figure}[h]
\centering
    \centering
    \includegraphics[width=\linewidth]{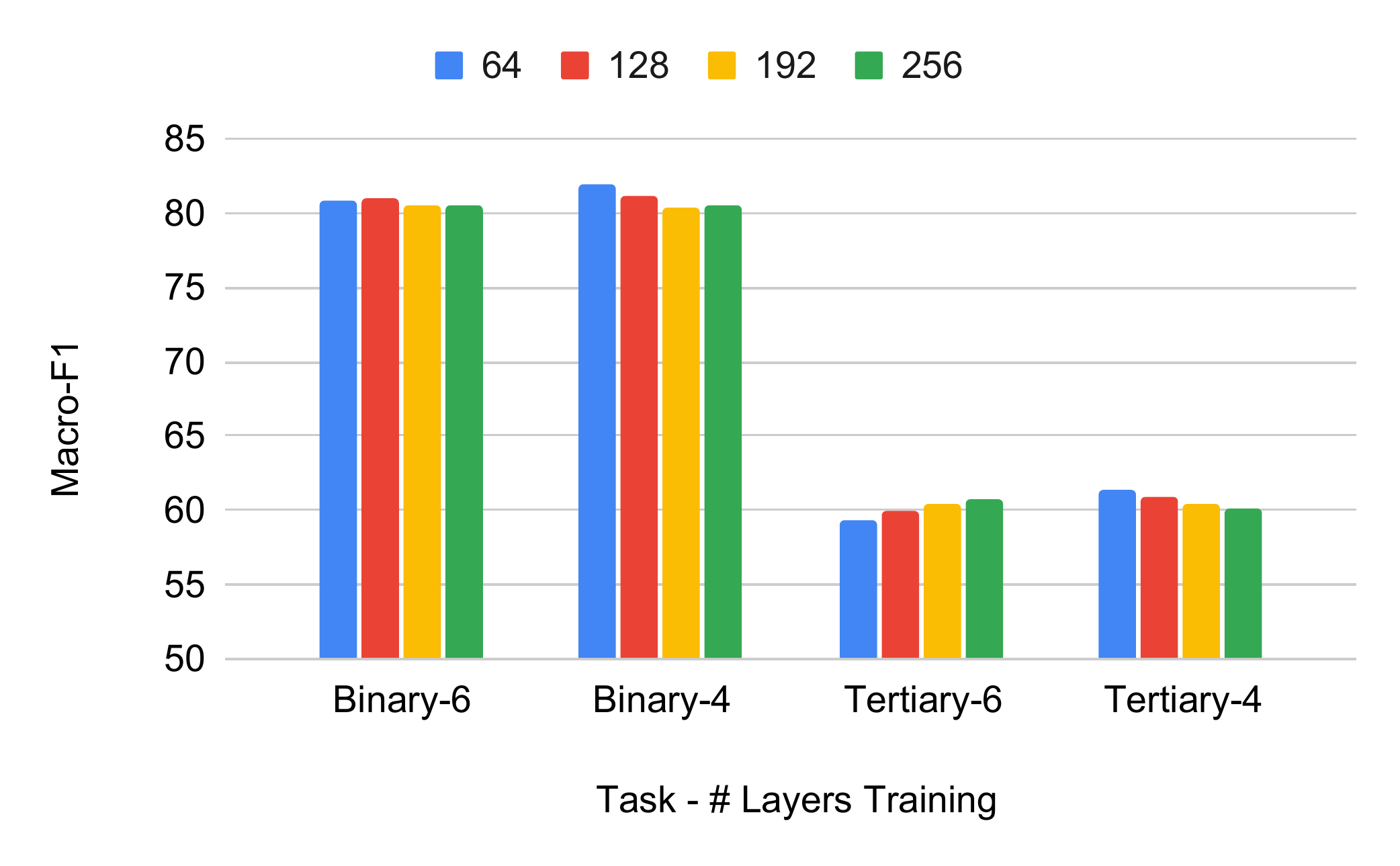}
    \caption{Ablation experiment on OCR text length (number of words) in \emph{ALBEF}}\label{fig:ocrlen}
\end{figure}

The amount of text that can be detected from an image varies, as it can be seen in Figure~\ref{fig:more_visual}. As a consequence, we experimented with the length of OCR text in terms of the number of words for both binary and tertiary claim detection with \emph{ALBEF}. We observe (see Figure~\ref{fig:ocrlen}) that 128 words give comparable or better performance than any less number of words in OCR text across tasks and number of layers fine-tuned. We chose 128 words instead of 64 because the model with 128 words showed a balanced performance for binary, tertiary and retrieved claims. Models with 64 or greater than 128 words had a lower performance for either visually-relevant or irrelevant retrieved claims.

\begin{figure}[h]
\centering
    \centering
    \includegraphics[width=\linewidth]{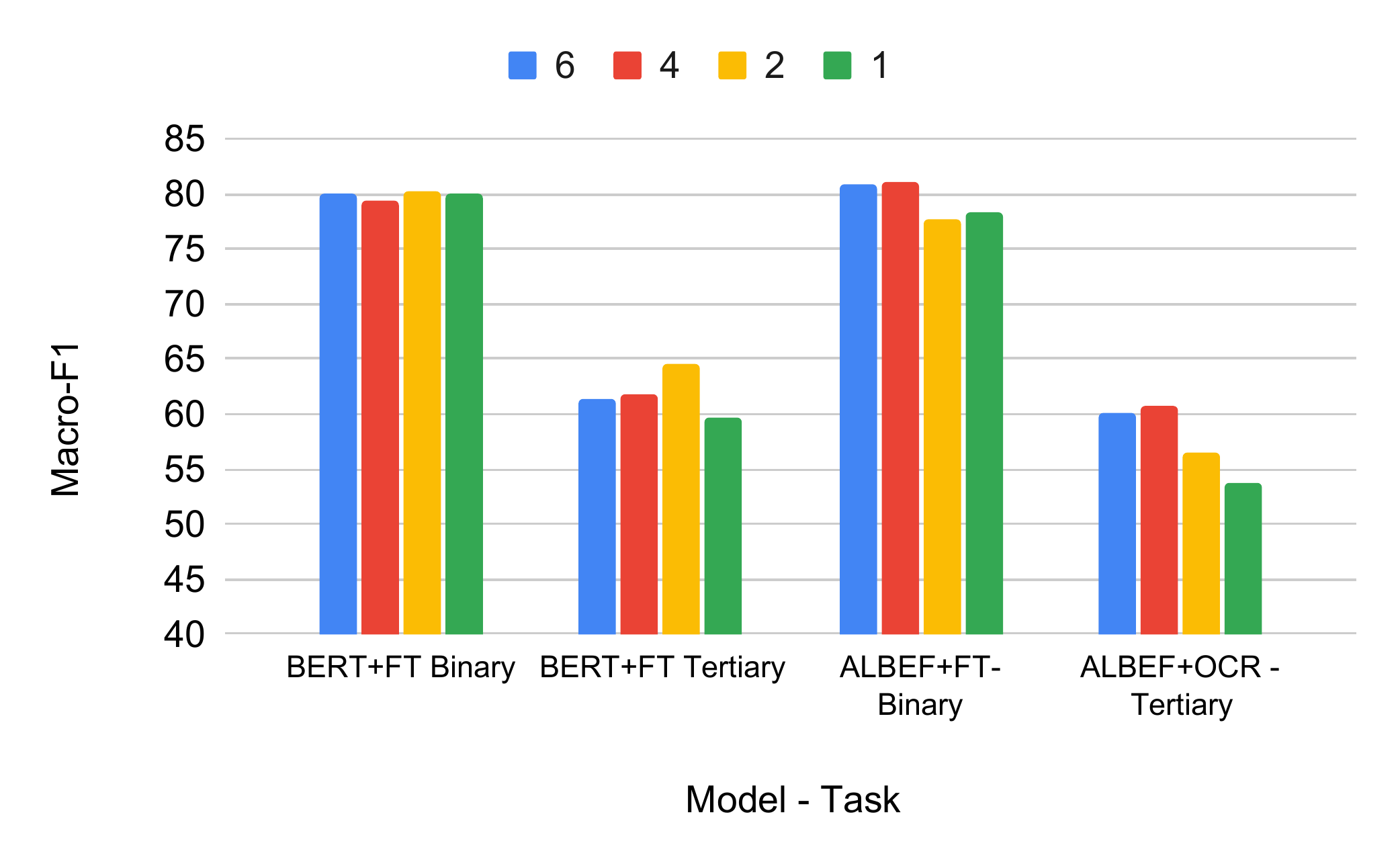}
    \caption{Ablation experiment on number of layers fine-tuned in \emph{BERT} and \emph{ALBEF}}\label{fig:layers}
\end{figure}

\subsubsection{Ablation on number of layers trained}\label{app:exp_num_layers}
We ran ablation experiments to see the effect of training the last few layers of \emph{BERT} and \emph{ALBEF} $\oplus$ OCR. We experiment with fine-tuning the last six, four, two layers and only the last layer of each model. The results are shown in Figure~\ref{fig:layers}. Overall, fine-tuning the last two and four layers of \emph{BERT} and \emph{ALBEF} respectively gives the best results. Therefore, all the fine-tuning results for \emph{BERT}, \emph{ALBEF} and \emph{ALBEF} $\oplus$ OCR are based on the above observation. For fine-tuning six or more layers, the unlabeled dataset can be incorporated in the future as a pre-training step followed by task-specific training.

\subsection{Additional Dataset and Annotation Details}\label{sec:add_data}

\subsubsection{Claim Definition}\label{app:claim_definition}

\noindent\textbf{Factually-verifiable Claims}: should ideally have some of the following information (extended from~\citet{barron2020overview}):
\begin{itemize}
\setlength\itemsep{0em}
    \item reference to who, where, when, what, etc
    \item a definition, procedure, law or a process
    \item numbers or quantities in the tweet, e.g. sums of money, number of cases or deaths
    \item verifiable predictions
    \item refers to people, events, (event) locations
    \item refers to images and videos in the tweet
    \item personal opinions with claims that have factually verifiable information
\end{itemize}

\noindent\textbf{Check-worthy Claims}: We follow a similar definition as %as followed by 
\citet{barron2020overview}, where claims are check-worthy if the information has some of the following properties:
\begin{itemize}
    \item \textit{Harmful}: if the statement attacks a person, organization, country, group, race, community, etc. The intention of such statements can be to spread rumours about an individual or a group, which should be checked by a professional or flagged and prioritized for further checking.
    \item \textit{Urgent or breaking news}: such statements are news-like where the claim is about prominent people (public personality like politicians, celebrities), organizations, countries and events (like disease outbreaks, forest fires, stock market crash).
    \item \textit{Up-to-date}: such claims often refer to official documents and contain parts of clauses in climate agreements or articles in a constitution. This information is vital for checking, as many people consume social media as means of news, information and believe it to be true.
\end{itemize}

\subsubsection{Filtering Strategies}\label{app:filtering_strategies}
The following Table~\ref{tab:stats} shows number of samples after each filtering step. The duplicate removal is performed across all the data irrespective of the topic in order to avoid duplicates that might fall into more than one topic.
\begin{table}[th]
\small % footnotesize
\centering
\begin{tabularx}{\linewidth}{|X|r|r|r|}
\hline
\textbf{Filtering Strategy}             & \textbf{COVID} & \textbf{Climate} & \textbf{Tech.}  \\
\hline
No Filter                     & \num{214715}         & \num{28374}            & \num{417403}              \\
\hline
Empty text                     & \num{214715}         & \num{28374}            & \num{417403}              \\
\hline
% \makecell[l]{Removal of \\ duplicates}          & 28522          & 11333            & 383043              \\
Duplicate removal          & \num{28522}          & \num{11333}            & \num{383043}              \\
\hline
Tweets with no image           & \num{28522}          & \num{11333}            & \num{383043}              \\
\hline
Text not in English            & \num{28148}          & \num{11274}            & \num{377532}              \\
\hline
Image size (200x200) & \num{27572}          & \num{10895}            & \num{369735}              \\
\hline
% Hashtags \textless 6           & \num{26786}          & \num{10013}            & \num{287242}              \\
Hashtags $>6$           & \num{26786}          & \num{10013}            & \num{287242}              \\
\hline
% \makecell[l]{Includes Hashtags \\ in Top-300}        & 17771          & 4874             & 62887               \\
Top-300 Hashtags        & \num{17771}          & \num{4874}             & \num{62887}               \\
\hline
\end{tabularx}
\caption{Data corpus statistics after applying different filtering strategies (in order).}\label{tab:stats}
\end{table}
% \egroup

\subsubsection{Class Distributions Across Topics}\label{app:class_distribution}

\begin{figure}[th]
    
    \centering
    \includegraphics[width=\linewidth,trim={0.5cm 0.6cm 0.5cm 0.6cm},clip]{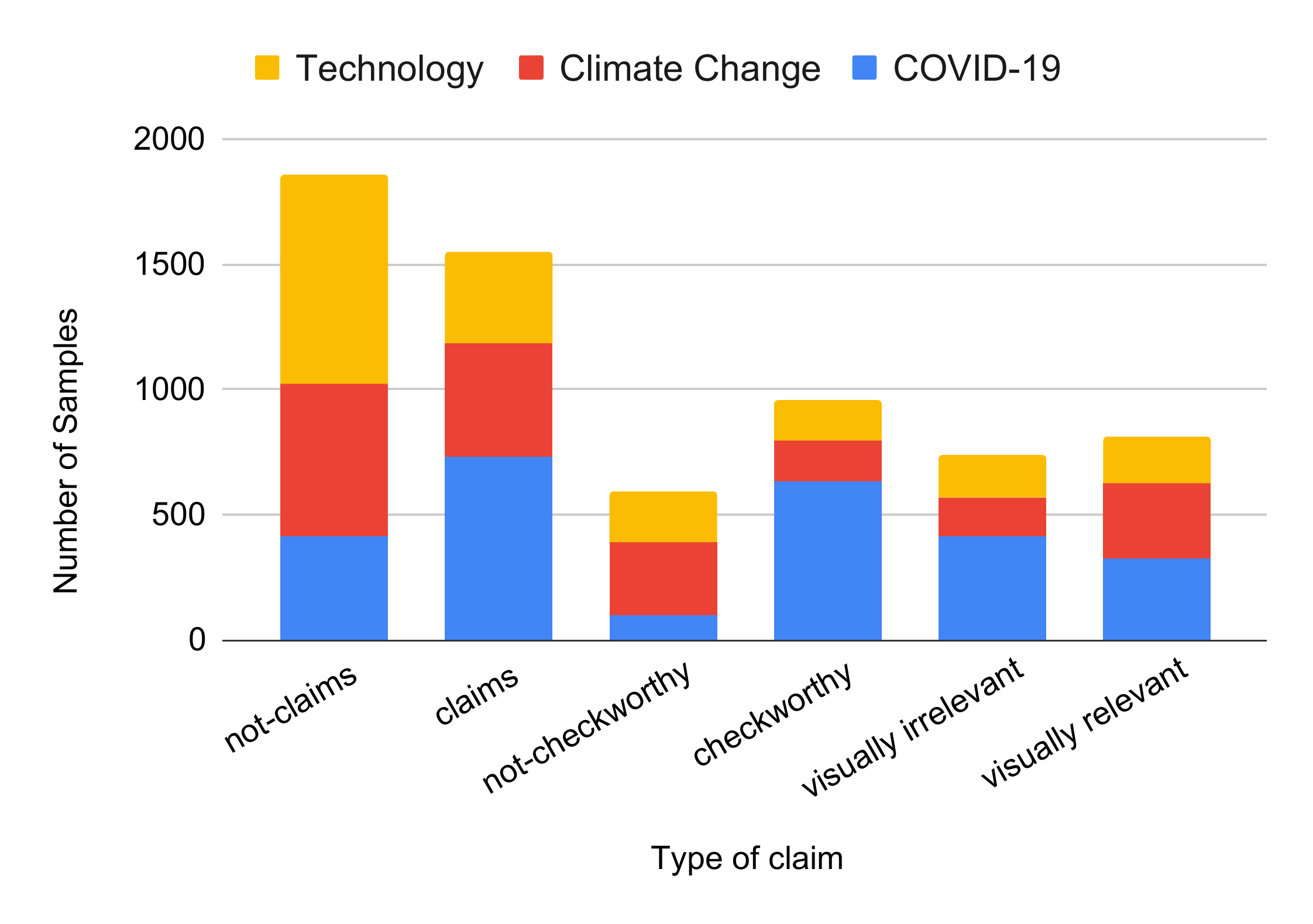}
    \caption{Class distributions in the annotated dataset ("with resolved conflicts") across different topics}\label{fig:class_stats}
\end{figure}

In Figure~\ref{fig:class_stats}, we provided the topic and class distributions in the labeled dataset.

\begin{table}[h]
	\small
	\centering
	\begin{tabularx}{\linewidth}{|X|X|X|X|}
		\hline
		\textbf{Types of Labels} & \textbf{COVID} & \textbf{Climate} & \textbf{Tech} \\
		\hline
		\textbf{Not Claims}            & \tr{306/34/73} 306/34/73   & \tr{449/38/120} 449/38/120           & \tr{617/81/136} 617/81/136        \\
		\hline
		\textbf{Claims}                & \tr{545/64/123} 478/58/104         & \tr{351/35/70} 251/24/48          & \tr{265/30/63} 198/21/44       \\
		\hline
		\hline
		\textbf{Not check-worthy}     & \tr{77/8/16} 25/4/3          & \tr{238/27/23} 141/16/5           & \tr{155/24/24} 97/9/8       \\
		\hline
		\textbf{check-worthy}          & \tr{468/56/107} 453/54/101         & \tr{113/8/47} 110/8/43           & \tr{110/6/39} 101/12/36        \\
		\hline
		\hline
		\textbf{Not Visual}            & \tr{302/31/78} 285/30/70        & \tr{112/8/33} 91/10/21           & \tr{125/15/34} 104/10/29       \\
		\hline
		\textbf{Visual}                & \tr{243/33/45} 193/28/34        & \tr{239/27/37} 160/14/27          & \tr{140/15/29} 94/11/15     \\
		\hline
		\hline
		\textbf{Total}                 & \tr{851/98/196} 784/92/177        & \tr{800/73/190} 700/62/168          & \tr{882/111/199} 815/102/180     \\
		\hline
	\end{tabularx}
	\caption{Labeled data characteristics in terms of type of labels and topic. Shown as Training/Validation/Testing splits. Second and third blocks are claims which are check-worthy (and not) and visual claims (and not) respectively. Red - "with resolved conflicts" and black - "without conflicts"}\label{tab:label_data}
\end{table}
% \egroup

\subsubsection{Conflict Resolution Strategy}\label{app:conflict_resolution}

Since three different users annotated each sample, a majority is always possible for the binary claim classification to derive unambiguous labels. However, a majority vote can not be achieved for the tertiary and visually-relevant claim classification task where all three annotators choose differently out of the possible three options. In Table~\ref{tab:conflict_table}, we provide the corresponding classes chosen by each annotator and the derived class after resolving the conflicts. The first case is resolved by giving priority to the \textit{claim but not check-worthy} label as \textit{check-worthiness} is a stricter constraint that is decided by only the majority. Two annotators indicated that the given sample is a claim (A-2 $\rightarrow$ Q1-Yes, A-3 $\rightarrow$ Q1-Yes). For the second case with visual claims, we select \textit{visually-relevant claim} label as there is a possibility of image and text being related even if one annotator marked "no" to the claim question (A-1 $\rightarrow$ Q1-No) but at least one annotator indicated that the sample is visually-relevant claim (A-3 $\rightarrow$ Q3-Yes).

\begin{table}[t]
\centering
\begin{tabular}{|l|l|l|l|}
\hline
\textbf{A-1} & \textbf{A-2} & \textbf{A-3} & \textbf{Derived Class} \\ \hline
Q1-No & \begin{tabular}[c]{@{}l@{}}Q1-Yes\\ Q2-No\end{tabular} & \begin{tabular}[c]{@{}l@{}}Q1-Yes\\ Q2-Yes\end{tabular} & \begin{tabular}[c]{@{}l@{}}\textit{Claim but not} \\ \textit{check-worthy}\end{tabular} \\ \hline
Q1-No & \begin{tabular}[c]{@{}l@{}}Q1-Yes\\ Q3-No\end{tabular} & \begin{tabular}[c]{@{}l@{}}Q1-Yes\\ Q3-Yes\end{tabular} & \begin{tabular}[c]{@{}l@{}}\textit{Visually} \\ \textit{relevant claim}\end{tabular} \\ \hline
\end{tabular}
\caption{Conflict resolution strategies to derive class labels where a majority vote can not be reached among three annotators (A) for check-worthiness and visual relevance tasks.}\label{tab:conflict_table}
\end{table}

\subsubsection{Split-wise Statistics}
The following Table~\ref{tab:label_data} shows split-wise distribution of topics and labels in data. Numbers in red and black are for "with resolved conflicts" and "without conflicts" splits, respectively. 
% \bgroup
% \setlength{\tabcolsep}{2pt} % Default value: 6pt
% \renewcommand{\arraystretch}{1} % Default value: 1

\subsubsection{Relevant Hashtags}: Although we crawl tweets from topic-based corpora, we further filter tweets by manually marking top 300 hashtags (sorted by occurrence) relevant to the topic. Figure~\ref{fig:htags} shows top-20 relevant hashtags for each topic.

\subsubsection{Annotation Tool} 
\begin{figure*}[]
\caption{Top-20 manually selected hashtags for topic relevance filtering strategy.}
\centering
\begin{subfigure}{0.32\linewidth}
    \caption{COVID-19}
    \centering
    \includegraphics[width=\linewidth]{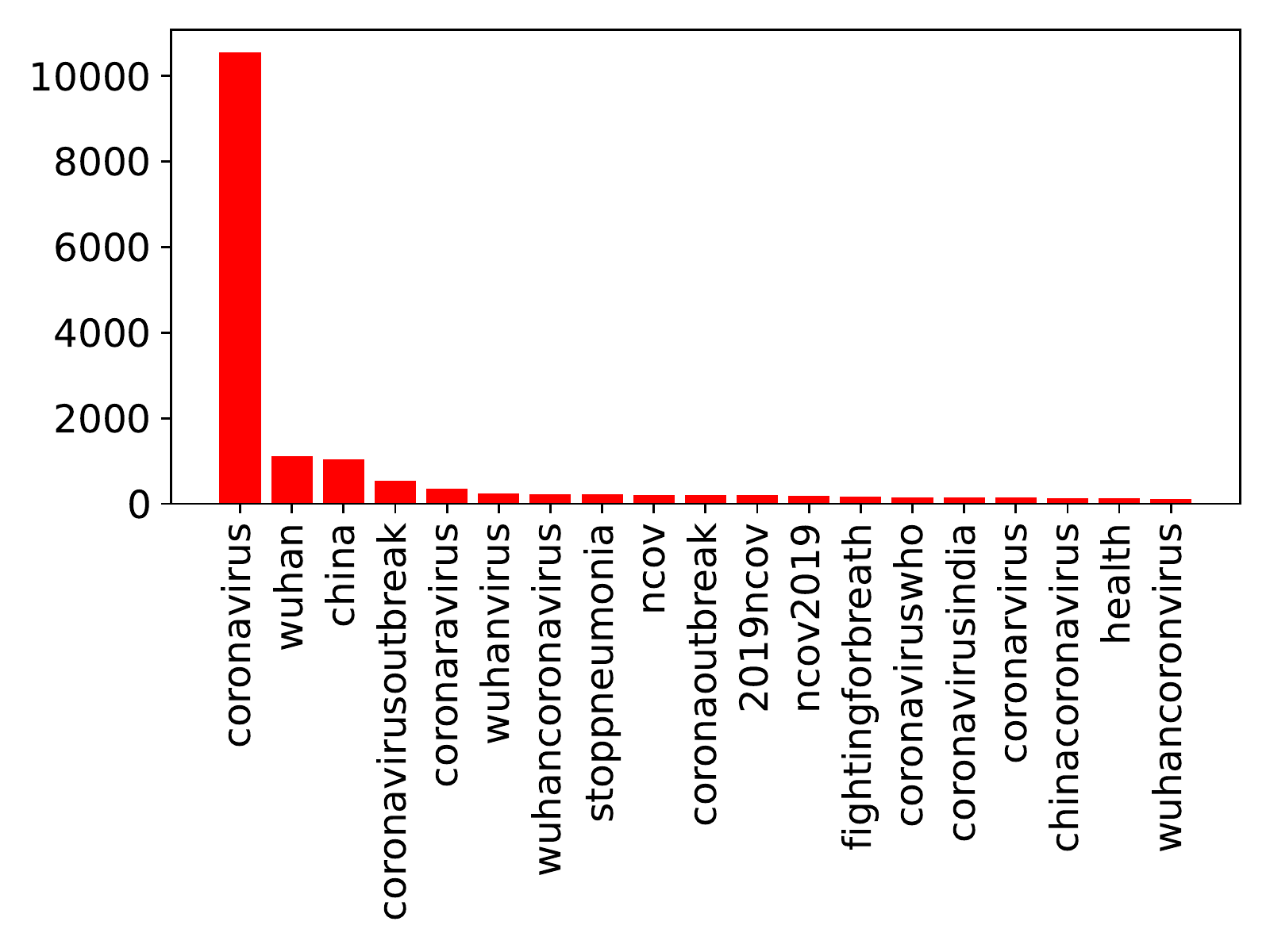}
\end{subfigure}
\begin{subfigure}{0.32\linewidth}
    \caption{Climate Change}
    \centering
    \includegraphics[width=\linewidth]{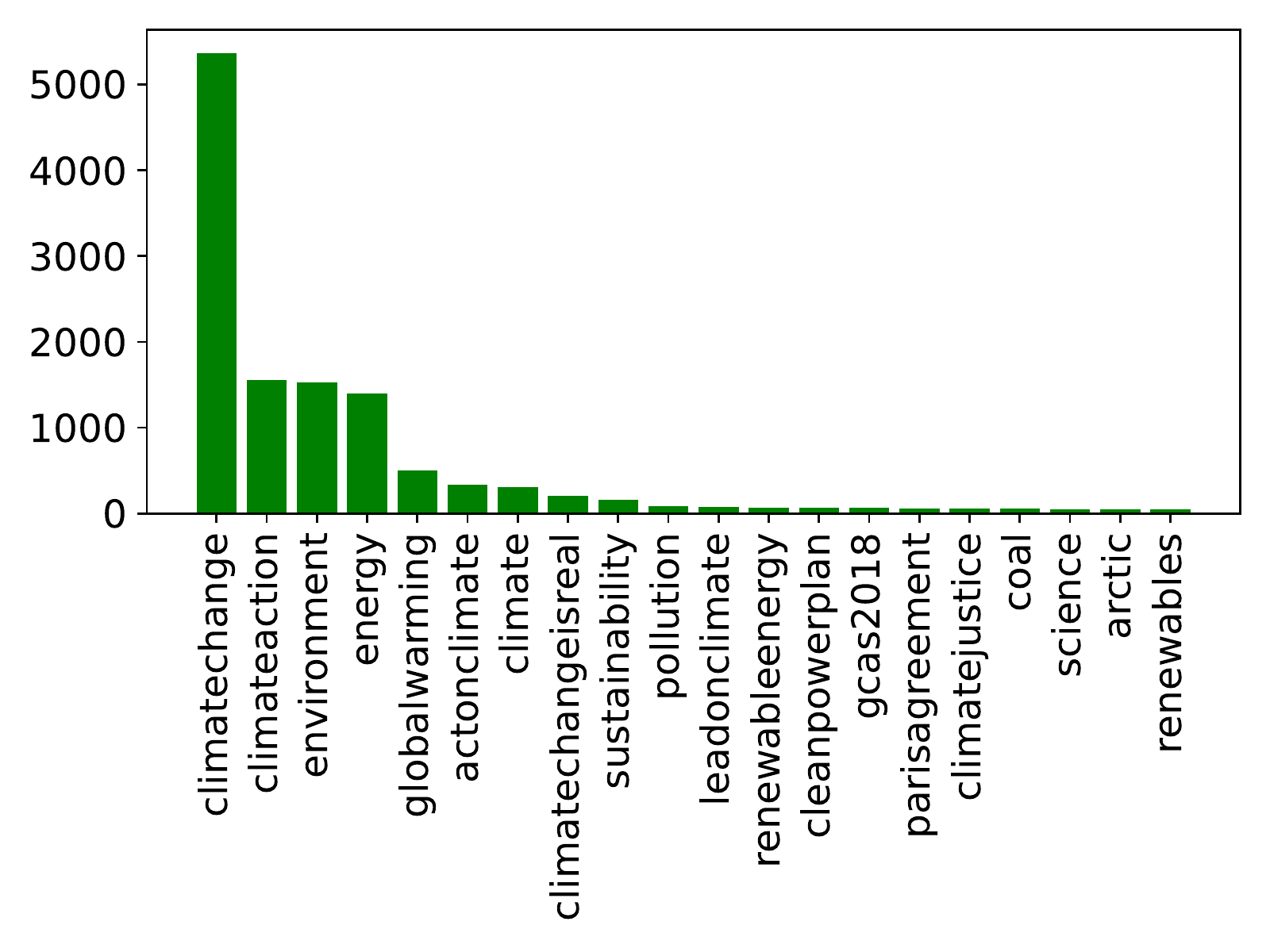}
\end{subfigure}
\begin{subfigure}{0.32\linewidth}
    \caption{Technology}
    \centering
    \includegraphics[width=\linewidth]{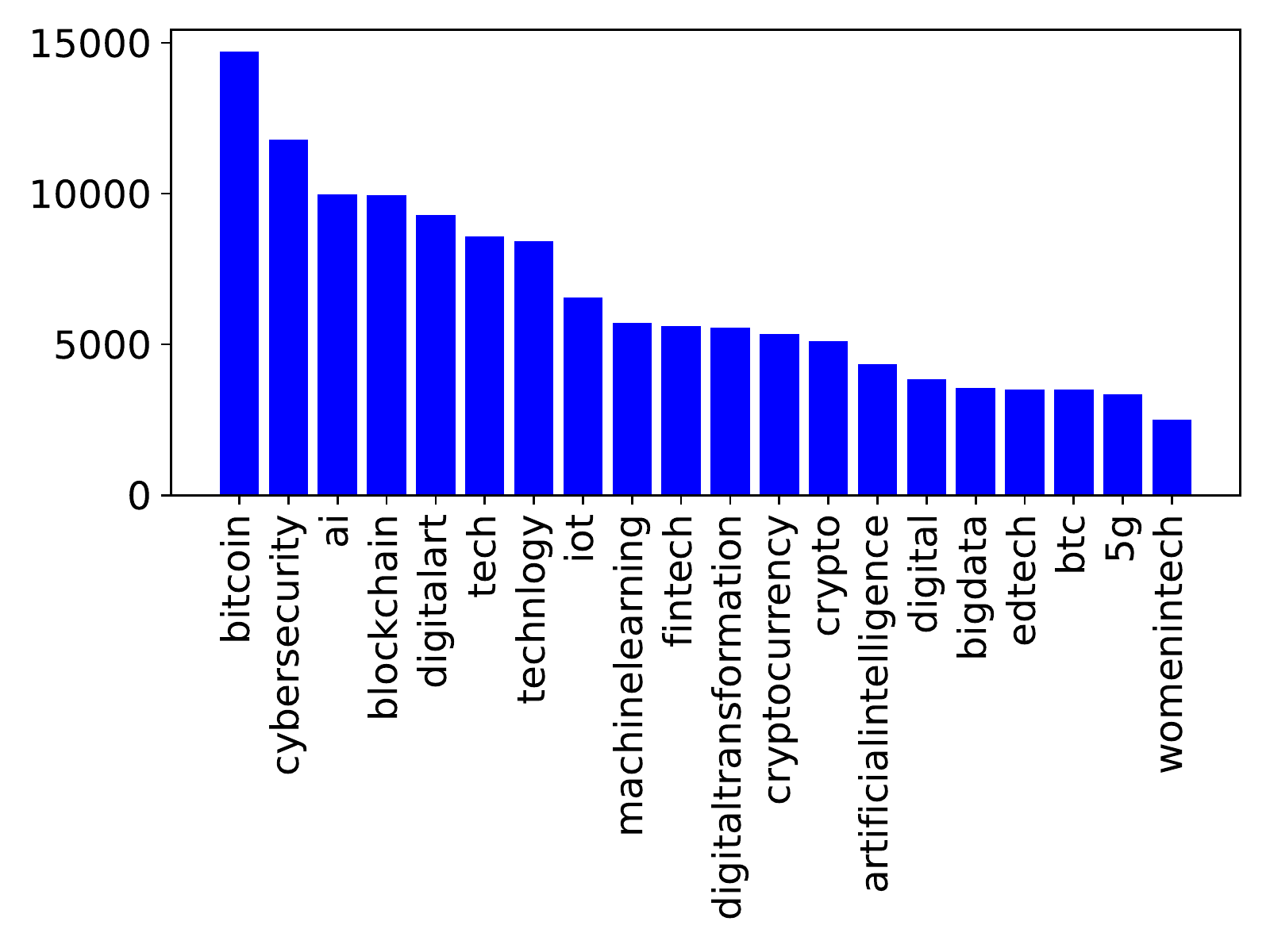}
\end{subfigure}
\\
\begin{subfigure}{\linewidth}
    \caption{Graphical User Interface that is used to annotate image-text tweets}
    \centering
    \includegraphics[width=\linewidth]{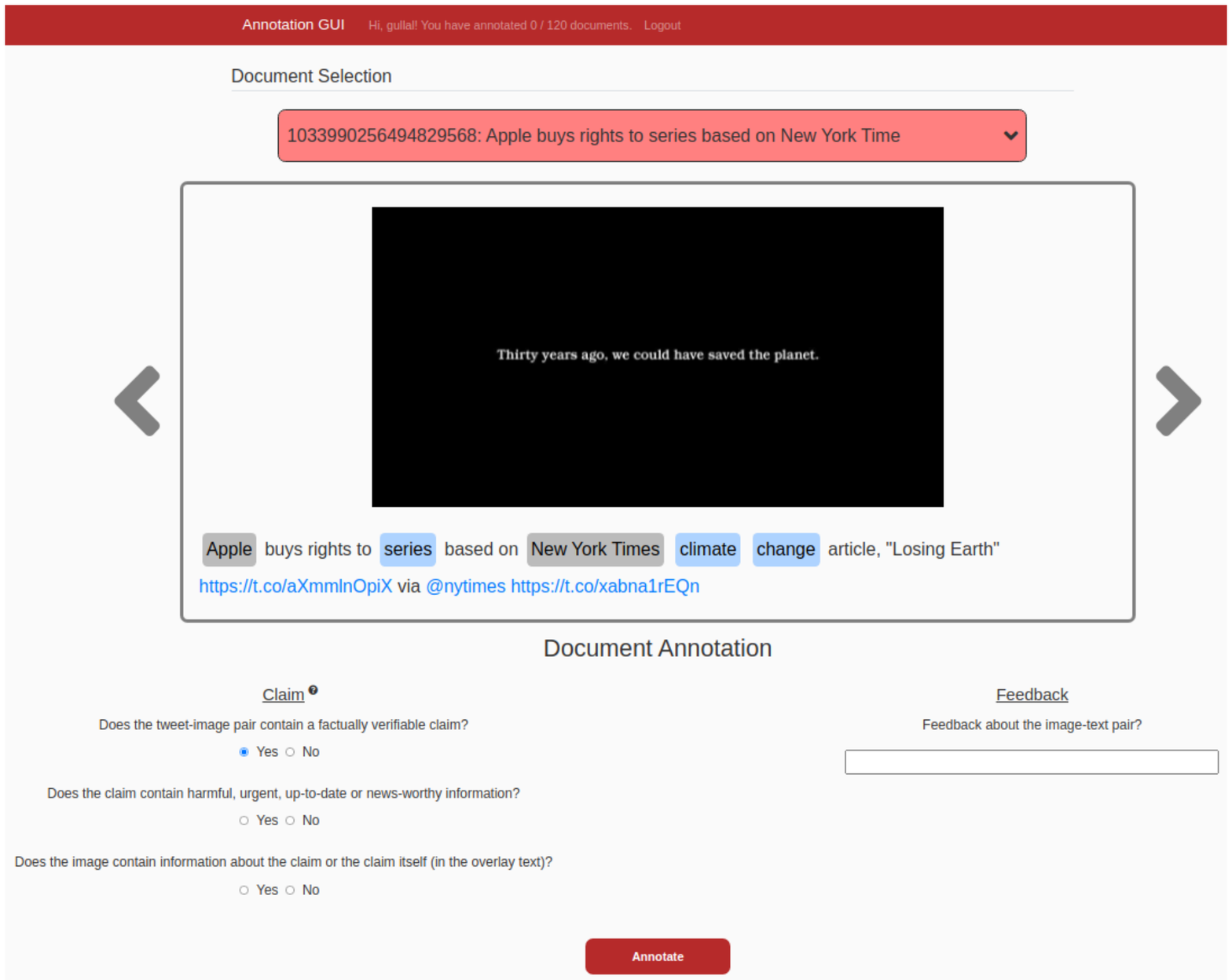}
    \label{fig:tool}
\end{subfigure}
\label{fig:htags}
\end{figure*}

Figure~\ref{fig:tool} shows the annotation screen with the image-text pair, claim questions and a text box for feedback on difficult and missing image tweets.

\subsection{Annotated Samples from the MM-Claims Dataset}\label{sec:add_samples}

\begin{figure}[]
% \centering
    \centering
    \includegraphics[width=\linewidth]{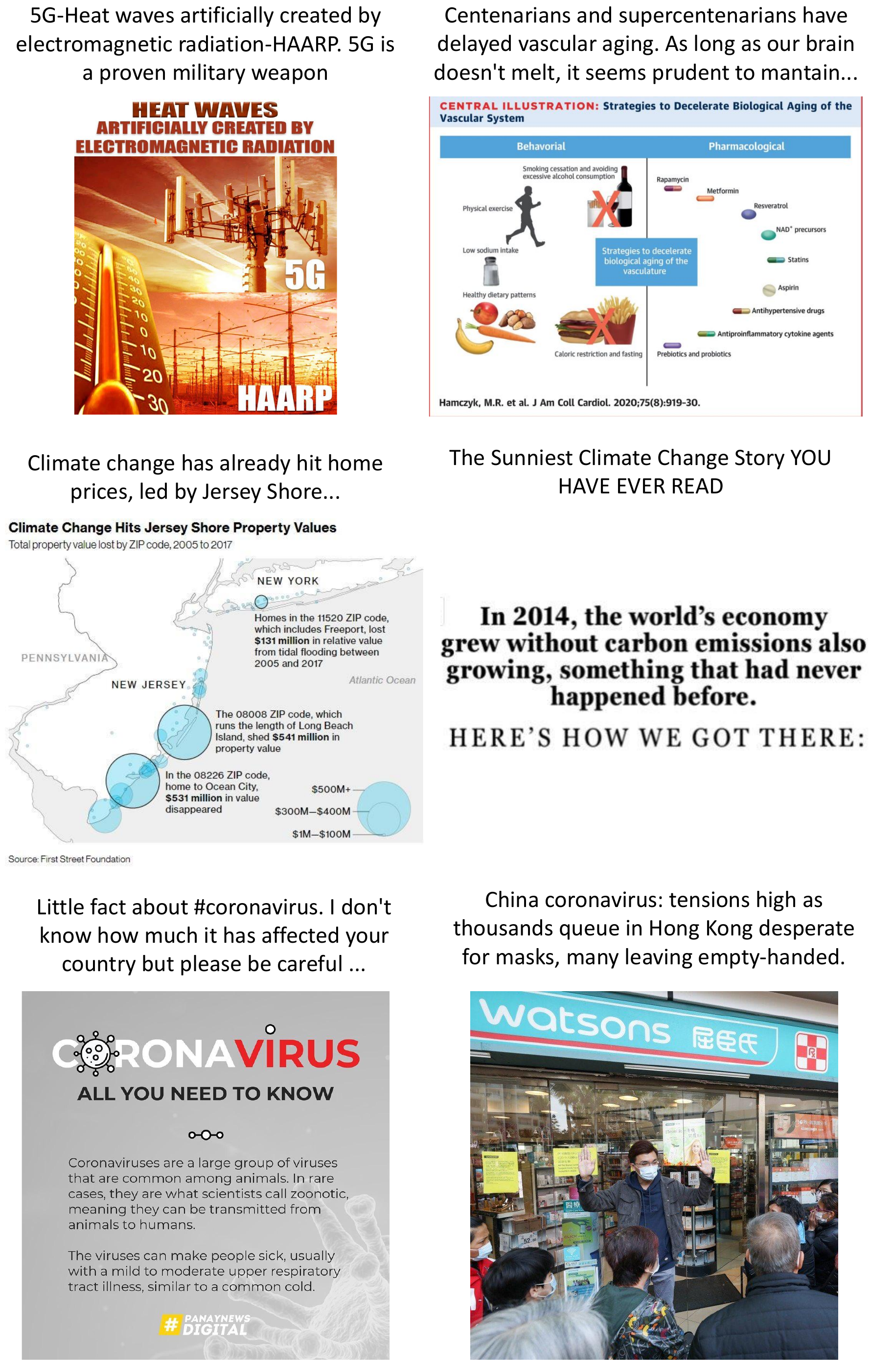}
    \caption{Additional examples for visually relevant claims for the topics \textit{COVID-19} (bottom row), \textit{Climate Change} (middle row), and \textit{Technology} (top row).}\label{fig:more_visual}
\end{figure}

\begin{figure}[b!]
% \centering
    \centering
    \includegraphics[width=\linewidth]{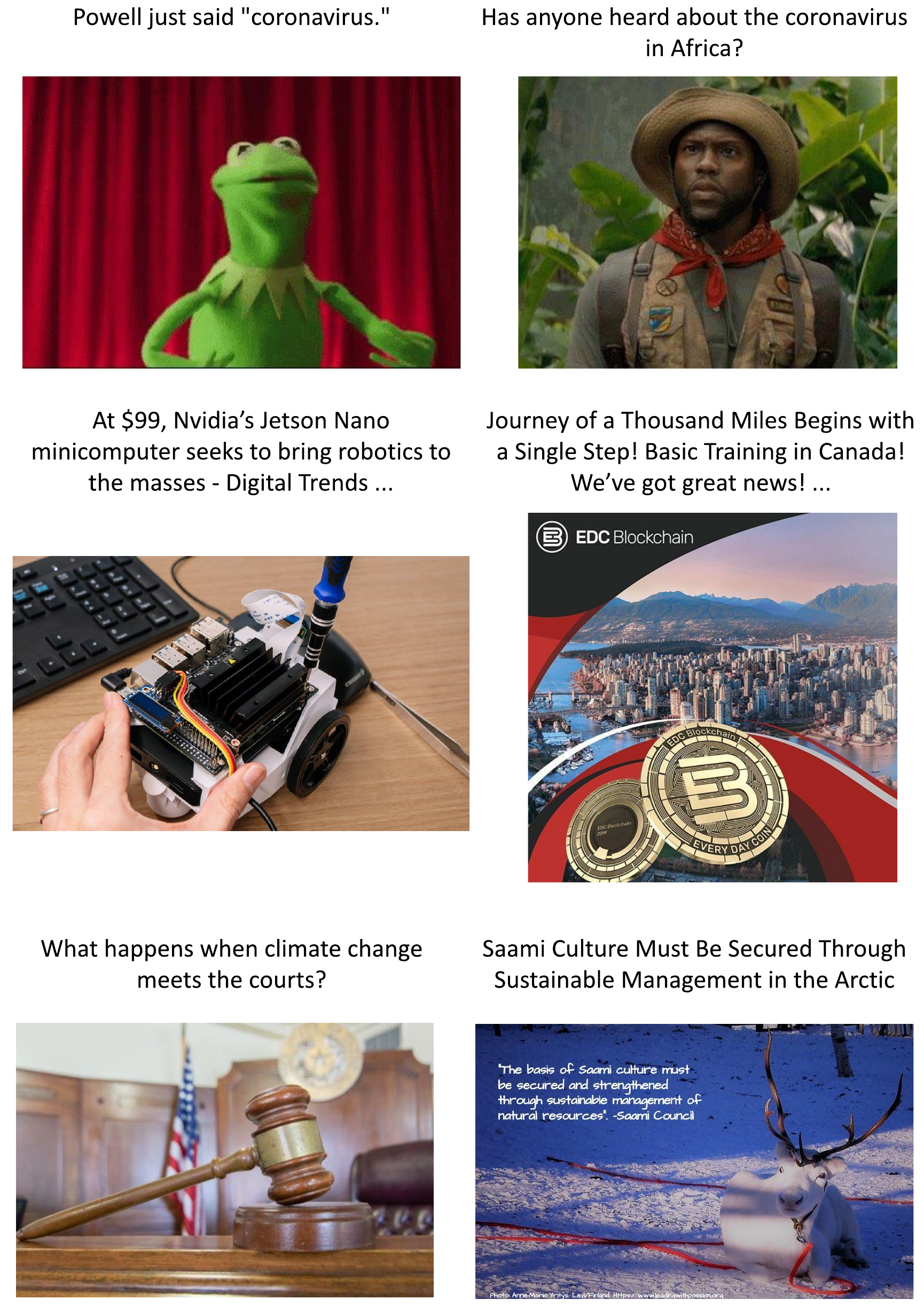}
    \caption{Additional examples that are not-claims for the topics \textit{COVID-19} (top row), \textit{Climate Change} (bottom row), and \textit{Technology} (middle row).}\label{fig:more_notclaims}
\end{figure}

We included multiple annotated samples corresponding to \textit{visually-relevant claim} (see Figure~\ref{fig:more_visual}) and \textit{not a claim} (see Figure~\ref{fig:more_notclaims}) classes.

% \begin{figure*}[]
% \caption{Annotated samples from MM-Claims dataset}

% \begin{subfigure}{0.45\linewidth}
%     \caption{Additional examples for visually relevant claims for the topics \textit{COVID-19} (bottom row), \textit{Climate Change} (middle row), and \textit{Technology} (top row).}
    
%     \includegraphics[width=\linewidth]{images/more_visual.pdf}
%     \label{fig:more_visual}
% \end{subfigure}
% \\
% \begin{subfigure}{0.45\linewidth}
%     \caption{Additional examples that are not-claims for the topics \textit{COVID-19} (top row), \textit{Climate Change} (bottom row), and \textit{Technology} (middle row).}
    
%     \includegraphics[width=\linewidth]{images/more_notclaims.pdf}
%     \label{fig:more_notclaims}
% \end{subfigure}
% \end{figure*}

\end{document}